\documentclass[letterpaper]{article} 
\usepackage{aaai2026}  
\usepackage{times}  
\usepackage{helvet}  
\usepackage{courier}  
\usepackage[hyphens]{url}  
\usepackage{graphicx} 
\usepackage{natbib}  
\usepackage{caption} 
\usepackage{algorithm}
\usepackage{algorithmic}

\usepackage{subfigure}
\usepackage{amsmath}
\usepackage{booktabs}
\usepackage{tcolorbox}
\usepackage{multirow}
\tcbuselibrary{listings}

\usepackage{newfloat}
\usepackage{listings}
\DeclareCaptionStyle{ruled}{labelfont=normalfont,labelsep=colon,strut=off} 
\DeclareFloatingEnvironment[fileext=lol,placement={tbp},name=Listing]{listing}
\title{Scalable and Accurate Graph Reasoning with LLM-Based Multi-Agents}
\author{
    Yuwei Hu\textsuperscript{\rm 1,2},
    Runlin Lei\textsuperscript{\rm 1,2},
    Xinyi Huang\textsuperscript{\rm 1},
    Zhewei Wei\textsuperscript{\rm 1}\thanks{Corresponding Author.},
    Yongchao Liu\textsuperscript{\rm 2}\thanks{Corresponding Author.}
}
\affiliations{
    \textsuperscript{\rm 1}Renmin University of China, Beijing, China\\
    \textsuperscript{\rm 2}Ant Group, Beijing, China\\
    huyuweiyisui@ruc.edu.cn, runlin\_lei@ruc.edu.cn, 2022201342@ruc.edu.cn,\\
    zhewei@ruc.edu.cn, yongchao.ly@antgroup.com
}

\usepackage{bibentry}

\begin{document}

\maketitle

\begin{abstract}
    Recent research has explored the use of Large Language Models (LLMs) for tackling complex graph reasoning tasks. However, due to the intricacies of graph structures and the inherent limitations of LLMs in handling long text, current approaches often fail to deliver satisfactory accuracy, even on small-scale graphs and simple tasks. To address these challenges, we introduce GraphAgent-Reasoner, a fine-tuning-free framework that utilizes a multi-agent collaboration strategy for explicit and precise graph reasoning. Inspired by distributed graph computation theory, our framework decomposes graph problems into smaller, node-centric tasks that are distributed among multiple agents. The agents collaborate to solve the overall problem, significantly reducing the amount of information and complexity handled by a single LLM, thus enhancing the accuracy of graph reasoning. By simply increasing the number of agents, GraphAgent-Reasoner can efficiently scale to accommodate larger graphs with over 1,000 nodes. Evaluated on the GraphInstruct dataset, our framework demonstrates impressive accuracy on polynomial-time graph reasoning tasks, significantly outperforming the best available models, both closed-source and fine-tuned open-source variants. Our framework also demonstrates the capability to handle real-world graph reasoning applications such as webpage importance analysis.
\end{abstract}


\section{Introduction}

Graphs, as a crucial data structure for modeling complex real-world relationships, are ubiquitous across various scenarios, \textit{e.g.} citation networks, recommendation networks. Many important applications like drug discovery~\citep{stokes2020deep}, traffic forecasting~\citep{jiang2022graph}, and financial detection~\citep{motie2023financial}, require reasoning over graphs to be realized.
Noticing the powerful general knowledge and language processing capabilities of Large Language Models (LLMs)~\citep{brown2020language}, a significant amount of works have focused on using LLMs to perform various reasoning tasks, such as mathematical formula derivation~\citep{meadows2023math}, commonsense reasoning~\citep{Madaan2022commonsense}, and multi-hop question answering~\citep{Creswell2023selection}. However, most of them primarily involve shallow or sequential reasoning. To bring the LLM reasoning closer to human thinking, it is necessary for LLMs to master deeper and more complex reasoning, such as graph reasoning.

Despite significant efforts by researchers to enable LLMs to memorize, comprehend, and perform basic reasoning on graph structures, several issues still persist: \textbf{1) The scale of graphs that can be handled is limited.} 
Describing graph structures in natural language inevitably leads to excessively long inputs. 
Due to context length limitations and the shortcomings of LLMs in handling lengthy text~\citep{Liu2023LostIT}, previous works~\citep{chai2023graphllm, fatemi2023talk, perozzi2024let} could only handle graphs of very limited size (e.g. fewer than 20 nodes and 100 edges). 
\textbf{2) The performance on graph reasoning tasks is relatively poor.} Unlike text, which can tolerate some degree of semantic deviation, reasoning and computation on graphs must be highly precise. However, current works demonstrate poor accuracy (average 20$ \sim $60\%) in various graph reasoning tasks like connectivity and shortest path.  
\textbf{3) Lacking explicit reasoning paths.}  Taking the shortest path as an example, the responses of existing models resemble a heuristic search approach to finding the shortest path on a graph, rather than strictly executing an algorithm. This makes it difficult to determine whether LLMs are genuinely deriving the answer through correct reasoning or merely making educated guesses. 
Although GraphWiz~\citep{chen2024graphwiz} attempts to generate explicit reasoning paths through fine-tuning, it often fails due to the presence of incomplete or wrong reasoning paths in its training data.
Furthermore, GraphWiz exhibits overfitting, which will be detailed in Section~\ref{case}.

\paragraph{Motivation.} The ultimate goal of graph reasoning is to enable LLMs to leverage graph-related knowledge or algorithms to solve real-world graph problems. However, with the development of information science and hardware storage, both the scale of graphs and the amount of information per node have become too large for a single LLM to handle. To address this, a natural idea is to use distributed approaches, where a large graph is stored across multiple LLMs separately and compute collaboratively. Therefore, just as graph algorithms have generally evolved from non-distributed to distributed forms~\citep{meng2024survey}), we hope that LLMs can also learn the concept of distributed processing, thereby harnessing the power of swarm intelligence to solve graph problems in real-world scenarios.

\begin{figure*}[htbp]
\centering
\includegraphics[width=0.7\textwidth]{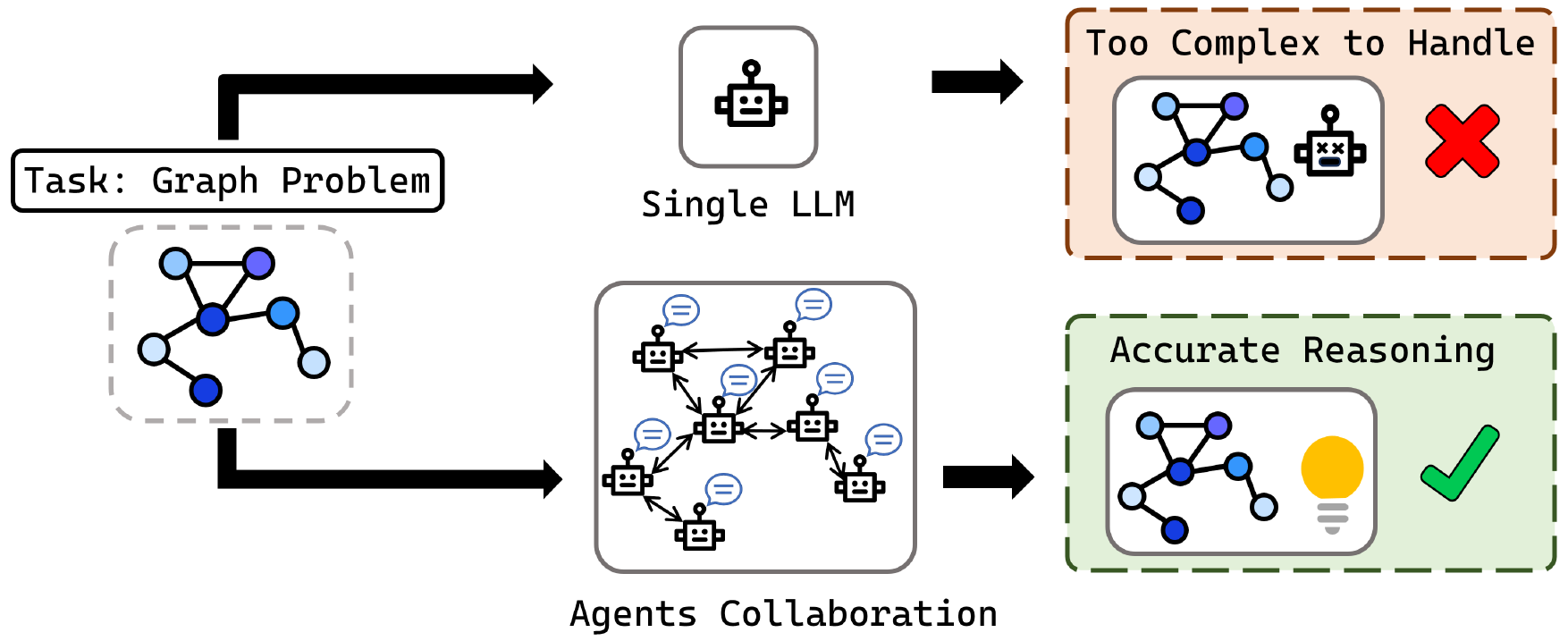}
\caption{The current situation of LLMs in solving graph problems. Previous methods using a single LLM often failed due to the complex graph structures. In contrast, our approach leverages agents collaboration to effectively address graph problems.}
\label{fig:intro}
\vspace{-0.6em}
\end{figure*}

\paragraph{Our Contribution.} To address the above limitations, in this paper, we propose the GraphAgent-Reasoner(GAR) framework, which leverages the power of swarm intelligence to solve graph reasoning problems, as shown in Figure~\ref{fig:intro}. We follow a node-centric approach, assigning an agent to each node, allowing it to focus on processing its own information and communicate with neighbors. 
Thus, we can easily scale up the size of graphs that can be processed by simply increasing the number of agents. 
At the same time, under the direction of a Master LLM, graph problems are decomposed into smaller, node-centric tasks, which are assigned to agents for collaborative resolution. 
This approach significantly reduces the scale and complexity of information each agent needs to process, thereby greatly improving the overall accuracy. 
Furthermore, since agents must clearly transmit the processed information to neighboring agents, the reasoning process becomes transparent, demonstrating the framework solves graph reasoning problems through clear and correct reasoning, rather than lucky guessing. In summary, our contributions are as follows:

\begin{itemize}
    \item We propose GraphAgent-Reasoner, a LLM-based multi-agents framework for graph reasoning, which requires no fine-tuning and can utilize any LLM as the backbone model. Our framework achieves impressive accuracy on various polynomial-time tasks, significantly surpassing the performance of existing methods.

    \item Our framework expands the scale of graph reasoning tasks handled by LLMs from 100 nodes to 1,000 nodes, demonstrating exceptional scalability. Furthermore, as the graph size increases, our framework does not exhibit the significant performance degradation seen in other methods and maintains robust accuracy.


    \item We explore the performance of our framework in real-world applications like webpage importance analysis, showcasing its potential for addressing complex graph reasoning problems in real-life situations.
\end{itemize}

\section{Preliminaries and Related Works}
\label{preliminaries}

\paragraph{Preliminaries.} In general scenarios, when discussing LLMs solving graph reasoning problems, the input is a ($\mathcal{G}$,$\mathcal{Q}$) pair. $\mathcal{G}$ is a graph represented as $\mathcal{G}=(\mathcal{V}, \mathcal{E}, \{s_i\}, \{t_i\})$, where $\mathcal{V}$ is the node set and $\mathcal{E}$, the edge set. For each node $v_i \in \mathcal{V}$, a sequential text node feature $s_i$ is associated; similarly, for each edge $e_i \in \mathcal{E}$, a sequential text edge feature $t_i$ is assigned. The graph $\mathcal{G}$ is described in natural language, typically using edge or adjacency list representation. $\mathcal{Q}$ is a task-specific instruction or problem description. LLMs will process the ($\mathcal{G}$,$\mathcal{Q}$) pair and return an answer string $A$. 

\paragraph{Large Language Models for Graph Reasoning.}
To further enhance the reasoning capabilities of LLMs, many works have attempted to improve the performance of LLMs in graph reasoning. \citet{wang2024can} first introduces the NLGraph Benchmark to evaluate the performance of LLMs on various graph reasoning tasks. \citet{fatemi2023talk} explores the impact of different graph encoding methods and graph structure types on the performance of LLMs in graph reasoning tasks. Additionally, it introduces another benchmark called GraphQA. Considering the lengthy nature of describing graph structures in text, \citet{chai2023graphllm} and \citet{perozzi2024let}  respectively use Transformers and GNNs to encode graph structures and attempt to align them with LLMs. Inspired by how humans understand structural information through the visual modality, \citet{Wei2024GITAGT} generates corresponding visual images based on graph structures and provides them to visual LLMs for graph reasoning. \citet{chen2024graphwiz} conducted Supervised Fine-Tuning and Directly Prefered Optimization on LLMs, enhancing the performance of LLMs and encouraging them to output explicit reasoning paths.

\paragraph{Large Language Model based Multi-Agents.}
Recent advancements in LLMs have spurred interest in their application within multi-agent systems. LLM-based multi-agent frameworks leverage the natural language understanding and reasoning capabilities of LLMs to enable agents to collaborate, communicate, and solve complex tasks in a distributed manner. Existing multi-agents works for problem solving primarily focuses on applications such as Software Development~\citep{dong2023self, hong2024metagpt, qian2024chatdev}, Embodied Agents~\citep{zhang2024build, zhao2024roco, chen2024scalable} and Science Debate~\citep{xiong2023examing, chan2024chateval}. However, using LLM-based multi-agents to handle graph data has been less explored, especially in the areas of graph reasoning and graph computation tasks. This may be due to the hallucination issue inherent in LLMs~\citep{huang2023hallucination}, where their responses are factually incorrect. This problem becomes more complex in a multi-agent setting, as the hallucinations of a single agent may propagate to other nodes by communication~\citep{guo2024multiagent}. This requires the performance of individual agents be sufficiently stable to ensure the correct operation of the entire multi-agent system.

\section{Limitations of a Single LLM }
Although LLMs exhibit strong language processing and logical reasoning capabilities, problems with the Transformer architecture and Attention mechanism~\citep{vaswani2017attention} still limit the scale and accuracy when they process graph problems. There are two primary limitations:

\begin{figure}[htbp]
\centering
\includegraphics[width=0.48\textwidth]{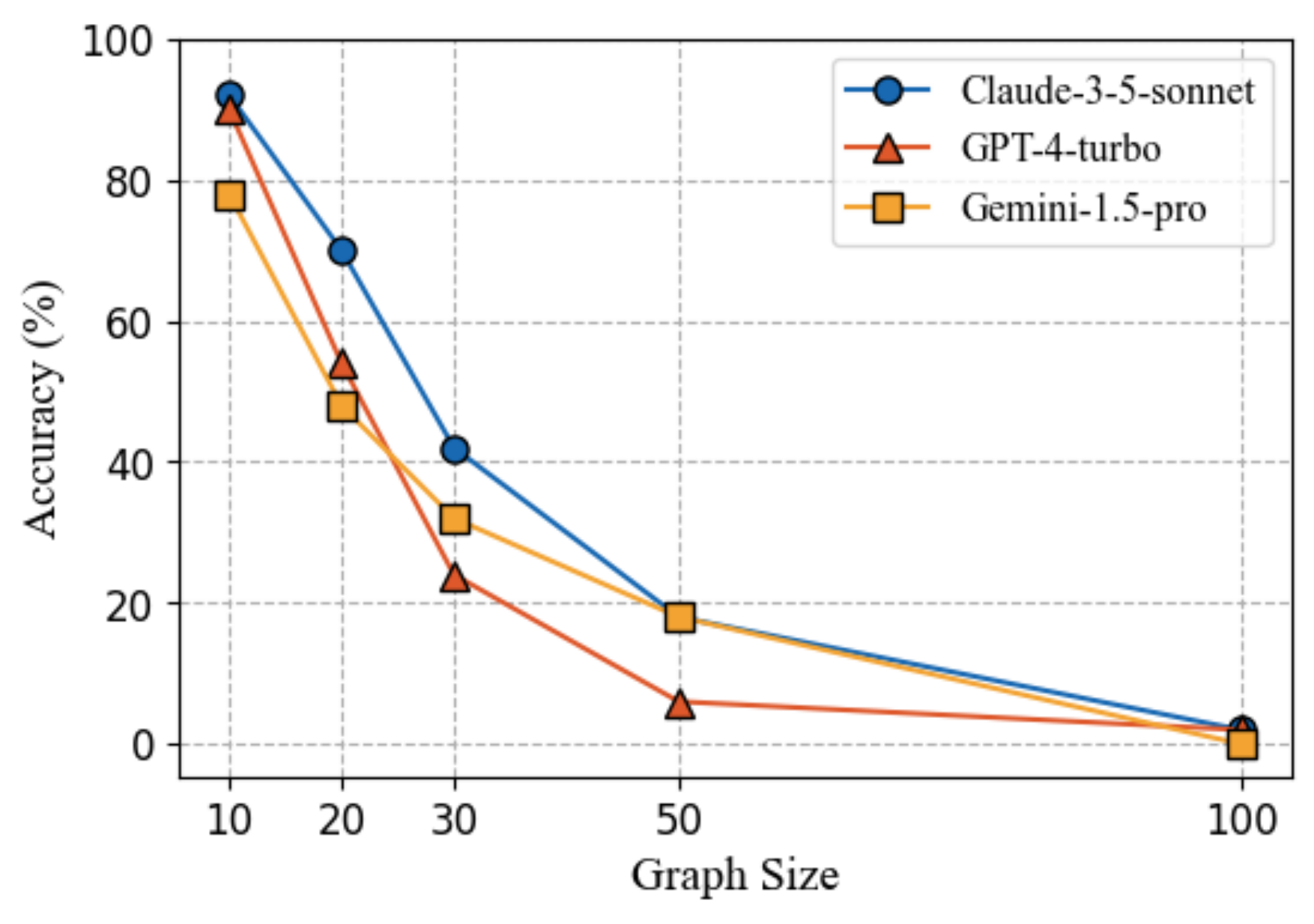}
\caption{The performance of a single LLM in memorizing first-order neighboring nodes. As the number of nodes increases, all models exhibit significant memory errors.}
\label{fig:remember}
\vspace{-1.0em}
\end{figure}

\paragraph{The graph structure is too complex to memorize and understand for a single LLM.} 
Using adjacency or edge lists to describe graph structures in natural language is the most intuitive and direct method, facilitating the processing of graph data by LLMs through text. However, this approach inevitably leads to a lengthy context, as the number of edges can grow quadratically with the number of nodes. 
As the graph scales up and becomes denser, the graph structure becomes highly complex, requiring a large amount of tokens to describe the edge relationships. When the text becomes too lengthy, it becomes difficult for LLMs to properly allocate attention, and they may even struggle with simple tasks such as key-value pair matching~\cite{Liu2023LostIT}. This presents significant challenges for LLMs in identifying key information for graph reasoning tasks from the lengthy context. Figure~\ref{fig:remember} shows the performance of a single LLM in memorizing one-hop neighbor nodes. We observe that as the number of nodes in the graph increases, various LLMs exhibit a significant decline in accuracy. 
If a single LLM cannot even correctly recall basic graph structural information like node neighbors, it becomes difficult to proceed with more complex graph reasoning or computation. 

Furthermore, the graph structure is described in a sequential manner. 
LLMs have to identify implicit graph structures from sequential text. Since the processing of LLMs is a black-box operation, it is difficult to assert that they truly construct graph structures implicitly and thereby understand them. \citet{huang2024can} conducted extensive experiments to explore whether LLMs treat the input prompts as graphs or merely as paragraphs with keywords on TAGs. The results show that the performance of LLMs in handling TAGs primarily stems from the context rather than the graph structure. LLMs tend to process the graph description as linearized paragraphs rather than graphs.


\paragraph{A single LLM struggles to solve reasoning problems in real-world scenarios.} 
Researchers fine-tune LLMs on graph reasoning tasks to empower themto enhance their ability to leverage learned graph-related knowledge or algorithms in solving real-world graph problems. However, in practical applications, the volume of information associated with each node can be substantial. Take citation networks as an example: a single node represents a paper, and its node information includes the title, abstract, and references, which could amount to several thousand tokens. In addition to the complexity of graph structures, the need to handle a large amount of node information further exacerbates the burden on a single LLM and highlights its shortcomings in processing long contexts. Moreover, using a single LLM to handle the entire network is inefficient, as it cannot coherently process the entire network's problems. Typically, it is necessary to manually compress or summarize the information for each node and then feed local subgraphs to the LLM for processing~\citep{guo2023gpt4graph, chen2024exploring}. 


Furthermore, many current works~\citep{chen2024graphwiz, perozzi2024let} require training GNNs or fine-tuning LLMs on individual or multiple graph reasoning tasks. However, when transferring to other graph tasks, a certain degree of performance degradation occurs, and retraining or fine-tuning for new graph tasks consumes a significant amount of time and resources. Whether LLMs can apply the graph knowledge and algorithms learned during the training process to actual graph reasoning also remains an open question. We explored this question in~\ref{case} and observed significant overfitting in LLMs fine-tuned on specific graph reasoning tasks. Therefore, the ideal solution would be to leverage the powerful general knowledge acquired during the pre-training phase of LLMs through an appropriate approach, enabling them to handle graph reasoning tasks as naturally as they do with natural language problems.

\begin{figure*}[!t]
    \centering
    \includegraphics[width=0.9\textwidth]{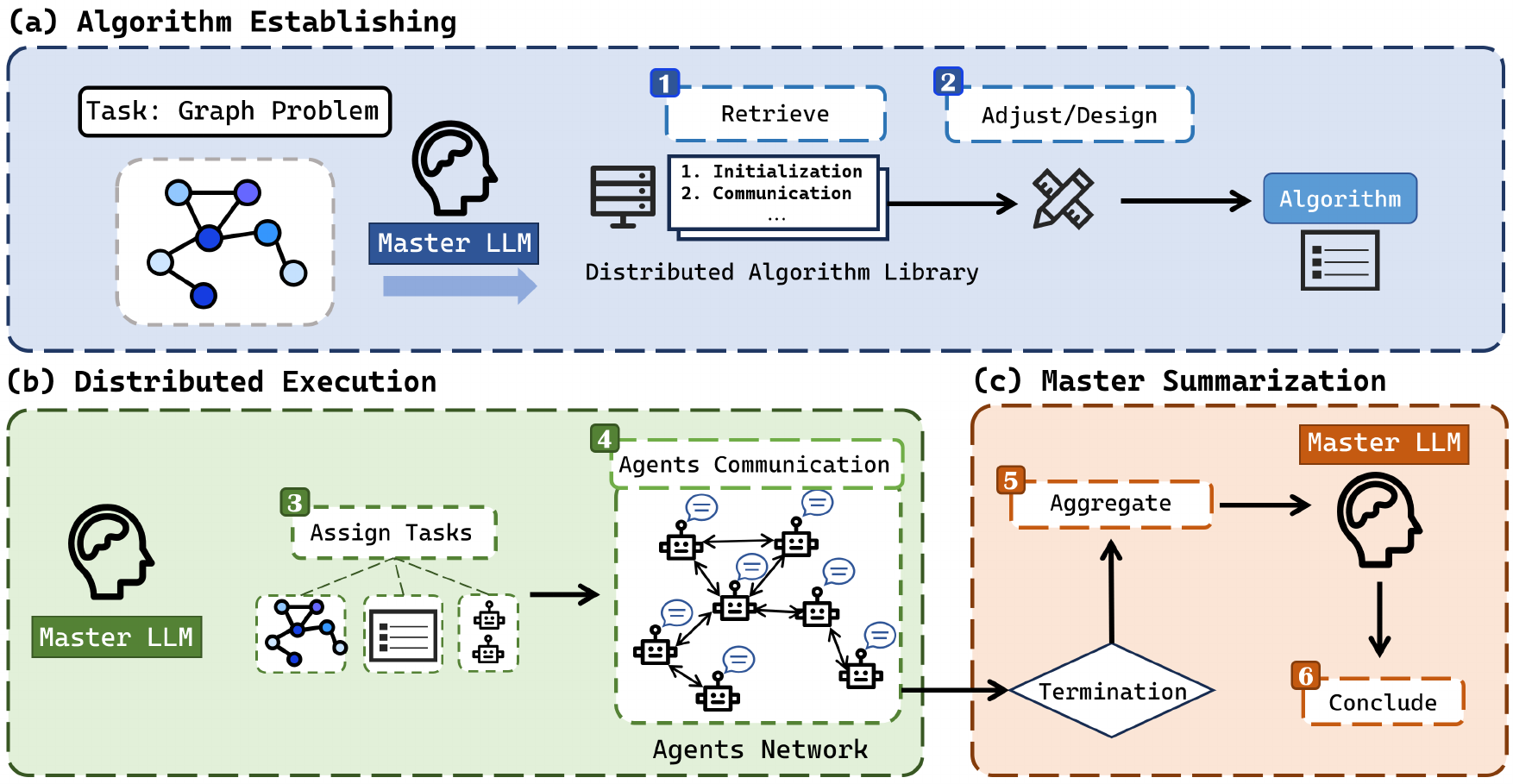}
    \caption{The framework of GraphAgent-Reasoner. Given a graph problem, the Master LLM will first construct agents network according to graph strcutures. It then sequentially performs Algorithm Establishing, Distributed Execution and Master Summarization, as detailed in this section.}
    \label{fig:framework}
    \vspace{-0.8em}
    \end{figure*}
    
\section{Our Approach}
To solve the limitations above, we propose a novel framework based on multi-agent collaboration called GraphAgent-Reasoner as shown in Figure~\ref{fig:framework}, aiming to solve graph reasoning problems explicitly and correctly. 
The interface of the framework is a powerful LLM called Master, which is responsible for processing the textual input of graph problems, constructing the agent network, directing them to collaboratively solve the problem, and finally aggregating the states of all agents to derive the solution. 
Its implementation is based on the React Agent proposed by \citet{yao2023react}, which is capable of reasoning based on the environment and executing corresponding actions, as detailed later. 
The pipeline of GAR consists of four steps: Graph Construction, Algorithm Establishing, Distributed Execution and Master Summarization.

\paragraph{Graph Construction.}
Given an input pair ($\mathcal{G}$, $\mathcal{Q}$), the Master first extracts the node and edge information from the textual description of graph $\mathcal{G}$. 
It then constructs an agent for each node and initializes the node’s state and neighbor information, forming an interconnected network of agents. 
Each agent independently maintains its state and neighbor data, communicates with adjacent agents based on instructions from the Master, and updates its state in each round.

\paragraph{Algorithm Establishing.} 
\label{algorithm_design}
To accommodate diverse graph tasks and fully exploit the knowledge embedded in LLMs during pre-training, we propose a unified solution approach framed within a distributed paradigm as shown in Algorithm~\ref{alg}. 
This approach requires the Master LLM to specify six core components for each problem: State, Message, Initialization, Send, Update, and Termination.

\begin{itemize}
    \item \textbf{State}: The local information maintained by each node, representing its current state. This can include attributes like node features, labels, long texts or any other task-specific data. The states evolve as nodes receive messages and update their information.
    \item \textbf{Message}: The information exchanged between neighboring nodes during communication rounds. Messages typically contain information that neighboring nodes need to perform updates, such as feature values, distances, or other task-relevant information.
    \item \textbf{Initialization}: At the start of the execution, each node initializes its state with predefined values, which may be based on node IDs, input features or task-specific requirements. This step ensures that the graph is ready to begin the communication process.
    \item \textbf{Send}: After initialization, each node generates messages based on its current state and sends them to its neighboring nodes. This step is repeated in each iteration, allowing nodes to continuously exchange information with their neighbors.
    \item \textbf{Update}: Upon receiving messages from its neighbors, each node updates its state by aggregating the incoming messages and combining them with its current state. This iterative process enables nodes to update their local information over time.
    \item \textbf{Termination}: The algorithm halts when a predefined stopping condition is met, such as reaching a fixed number of iterations, achieving convergence, or satisfying a task-specific criterion. Once the termination condition is reached, each node will send its final state to the Master LLM, and the execution terminates.
\end{itemize}

\begin{algorithm}[h]
\caption{Distributed Paradigm}
\label{alg}
\textbf{Input}: Agent Nodes $\mathcal{A}$, each agent $a \in \mathcal{A}$ maintains a state $S_a$, the maximum iterations $I_{max}$ given by the Master LLM\\
\textbf{Output}: Final state $S_a$ for each agent $a \in \mathcal{A}$
\begin{algorithmic}[1]
\item[\hspace{\algorithmicindent}] /* \textit{Initialization} */
\STATE Each agent $a \in \mathcal{A}$ initializes its state $S_a$ based on \textbf{Initialization} rules.
\STATE Each agent $a$ sends an initial message $M_{a \rightarrow v}$ to each of its neighbors $v \in \text{Neighbors}(a)$ based on its current state $S_a$ and \textbf{Send} rules.
\item[\hspace{\algorithmicindent}] /* \textit{Communication} */
\WHILE{\textit{Iteration} $i < I_{max}$ \textit{and} \textbf{Termination} \textit{not met}}
\item[\hspace{\algorithmicindent}] /* \textit{Receive} */
\STATE Each agent $a$ receives messages $M_{u \rightarrow a}$ from all neighboring agents $u$.
\item[\hspace{\algorithmicindent}] /* \textit{Update} */
\STATE Each agent $a$ updates its state $S_a$ based on the received messages $M$ and its own current state $S_a$ according to \textbf{Update} rules.
\item[\hspace{\algorithmicindent}] /* \textit{Send} */
\STATE Each agent $a$ sends updated messages $M_{a \rightarrow v}$ to each of its neighbors $v$ based on the updated state $S_a$ according to \textbf{Send} rules.
\ENDWHILE
\STATE \textbf{return} the final state $S_a$ for all agents $a \in \mathcal{A}$
\end{algorithmic}
\end{algorithm}

Since LLMs lack prior knowledge of this distributed paradigm, to facilitate the Master LLM's understanding and application of the framework, we develop a distributed algorithm library that adheres to this distributed paradigm, from which the Master LLM can query relevant algorithm templates to generate distributed solutions within this paradigm. Specifically, we selected classic distributed graph algorithms and documented their implementations under this distributed paradigm. Some examples are presented in Appendix~\ref{app:distributed_in}. Drawing on prior work~\citep{zheng2024executing, jojic2024gpt}, we endeavor to write detailed reasoning steps of each part in the algorithm to encourage the agent to think step by step as much as possible, which plays an important role in enhancing the success rate of individual agents.  

When receiving a problem input, the Master LLM first retrieves the $k$ algorithms most relevant to the problem description from the distributed algorithm library. If there are algorithms suitable for handling the problem, the Master LLM will adjust the algorithm according to the problem description, such as changing the initialization and termination conditions (e.g., the source node in the shortest path problem). If there are no appropriate algorithms, the Master LLM will design a distributed algorithm following the distributed paradigm based on the examples of the retrieved algorithms. For some generated examples, see Appendix~\ref{app:distributed_out}.

\paragraph{Distributed Execution.} After the distributed algorithm is designed, the Master LLM will relay the approach to each agent node for execution according to the process outlined in Algorithm~\ref{alg}. Each agent will first initialize its state based on node information and algorithm rules and then send an initial message to neighboring agents. Subsequently, each agent will iteratively execute the operations of receiving messages, updating its state, and sending messages according to the algorithm rules, synchronizing progress after each communication round. 
Communication will continue until the maximum number of iterations is reached or the termination condition is met.

\paragraph{Master Summarization.} Finally, the final state of all agent nodes will be aggregated to the Master LLM, which will summarize the results conclude based on the problem and return the final answer in natural language form.
\section{Experiments}
In this section, we summarize the key experiments conducted with GAR. We begin by highlighting some of the most exciting results from our analysis here:

\begin{itemize}
    \item \textbf{R1}: GAR achieves impressive accuracy on polynomial-time graph reasoning problems, significantly surpassing existing closed-source models and open-source models fine-tuned on extensive data.
    \item \textbf{R2}: GAR maintains high accuracy on larger-scale graphs (\textbf{up to 1000 nodes}), demonstrating superior scalability. 
    In contrast, as the number of nodes increases, other models exhibit a significant decline in performance or become incapable of handling the problem at all due to the context length limitation.
    \item \textbf{R3}: GAR showcases a robust understanding and application of graph algorithms in real-world graph reasoning scenarios, highlighting its potential for addressing complex graph problems encountered in daily life. In contrast, other open-source models that have undergone extensive fine-tuning on graph reasoning datasets fail to apply the learned graph reasoning knowledge when confronted with rephrased real-world graph problems.
\end{itemize}

\begin{table*}[!t]
    \centering
    \caption{Performance of GraphAgent-Reasoner and other models on polynomial-time tasks of GraphInstruct test set. Each task contains 400 test cases, with a maximum of 100 nodes. The first best result for each task is highlighted in bold, and the second best result is highlighted underlined.}
    \resizebox{0.9\textwidth}{!}{%
    \begin{tabular}{lccccccc}
    \toprule
    \multirow{2}{*}{\textbf{Models}} & \multicolumn{4}{c}{\textbf{Linear}} & \multicolumn{2}{c}{\textbf{Polynomial}} & \multirow{2}{*}{{\textbf{Average}}} \\
    \cmidrule(lr){2-5} \cmidrule(lr){6-7}
    & cycle & connect & bipartite & topology & shortest & triangle &\\
    \midrule
    \textbf{Closed-source Models} & & & & & & &\\
    GPT-4 (zero-shot) & 38.75 & 17.00 & 65.25 & 5.00 & 9.25 & 5.75 & 23.50 \\
    GhatGPT (2-shot) & 51.25 & 43.75 & 70.75 & 4.50 & 3.50 & 17.25 & 31.83 \\
    GPT-4 (2-shot) & 52.50 & 62.75 & 74.25 & 25.25 & 18.25 & 31.00 & 44.00 \\
    \midrule
    \textbf{Fine-tuned Open-source Models} & & & & & & & \\
    Naive SFT (LLaMA 2-7B) & 73.75 & 83.50 & 41.25 & 4.00 & 9.50 & 30.00 & 40.17 \\
    Naive SFT (Mistral-7B) & 73.75 & 83.50 & 78.50 & 1.00 & 23.00 & 47.00  & 	51.13 \\
    GraphWiz (LLaMA 2-7B) & 91.50 & 87.00 & 74.00 & 18.00 & 28.00 & 38.25 & 56.13\\
    GraphWiz (Mistral-7B) & \underline{92.00} & \underline{89.50} & 72.00 & 19.00 & \underline{31.25} & 38.75 & 57.08 \\
    GraphWiz-DPO (LLaMA 2-7B) & 89.00 & 82.50 & 84.75 & 46.75 & 24.00 & \underline{52.75} & \underline{63.29} \\
    GraphWiz-DPO (Mistral-7B) & 85.50 & 79.50 & \underline{85.50} & \underline{85.25} & 12.50 & 29.00 & 62.88 \\
    \midrule
    GraphAgent-Reasoner & \textbf{99.50} & \textbf{100.00} & \textbf{100.00} & \textbf{96.50} & \textbf{99.75} & \textbf{93.25} &  \textbf{98.00} \\
    \bottomrule
    \end{tabular}%
    }
    \label{performance}
\end{table*}

\paragraph{Datasets.} We conduct our experiments on the graph reasoning tasks proposed in GraphInstruct~\citep{chen2024graphwiz}. This dataset contains nine graph reasoning problems with different time complexity, ranging from linear and polynomial complexity to NP-complete.

\begin{itemize}
    \item \textbf{Linear.} Cycle Detection (Detect if a given graph $\mathcal{G}$ contains any cycles), Connectivity (Assess if two nodes $u$ and $v$ in a given graph $\mathcal{G}$ are connected via a path), Bipartite Graph Check (Judge if a given graph $\mathcal{G}$ is bipartite), and Topological Sort (Find a topological ordering of vertices in a directed acyclic graph $\mathcal{G}$).
    \item \textbf{Polynomial.} Shortest Path (Compute the shortest path between two specific nodes $u$ and $v$ in a given graph $\mathcal{G}$), Maximum Triangle Sum (Find the maximum sum of weights for any connected triplet of vertices in a given graph $\mathcal{G}$), and Maximum Flow (Calculate the maximum flow from a source node $s$ to a sink node $t$ in a directed graph $\mathcal{G}$).
\end{itemize}
Due to the complexity of NP-complete problems, there are currently no mature exact distributed algorithms available for their solution. Consequently, the Master LLM is unable to design correct and effective distributed algorithms based on the knowledge acquired during pre-training. Therefore, in our experiments, we only consider linear and polynomial-time problems. Detailed information of the dataset and partial test results for NP-complete problems will be presented in Appendix~\ref{app:dataset}. 

\paragraph{Setting.}  The underlying reasoning LLM of Agent Node used in our framework is ChatGPT-4o-mini-2024-07-18, and the base model of Master LLM is ChatGPT-4-turbo\citep{achiam2023gpt}. The temperature is consistently set to 0. Our framwork is built upon AgentScope~(\cite{agentscope}), an innovative platform to easily build reliable, high-performance multi-agent applications.

\subsection{Performance on GraphInstruct}

In this experiment, we evaluate the performance of GAR on polynomial-time tasks of the GraphInstruct dataset. The results are shown in Table~\ref{performance}.
We see GAR exhibits impressive results on these tasks, significantly outperforming other models. Especially on shortest and triangle tasks with high time complexity, GAR substantially improves the performance of LLMs. Problems that a single LLM struggles to solve have been effectively resolved through collaboration by agents after being decomposed into smaller, node-centric tasks.

\begin{figure}[htbp]
    \centering
    \subfigure[Cycle Detection]{
        \includegraphics[width=0.21\textwidth]{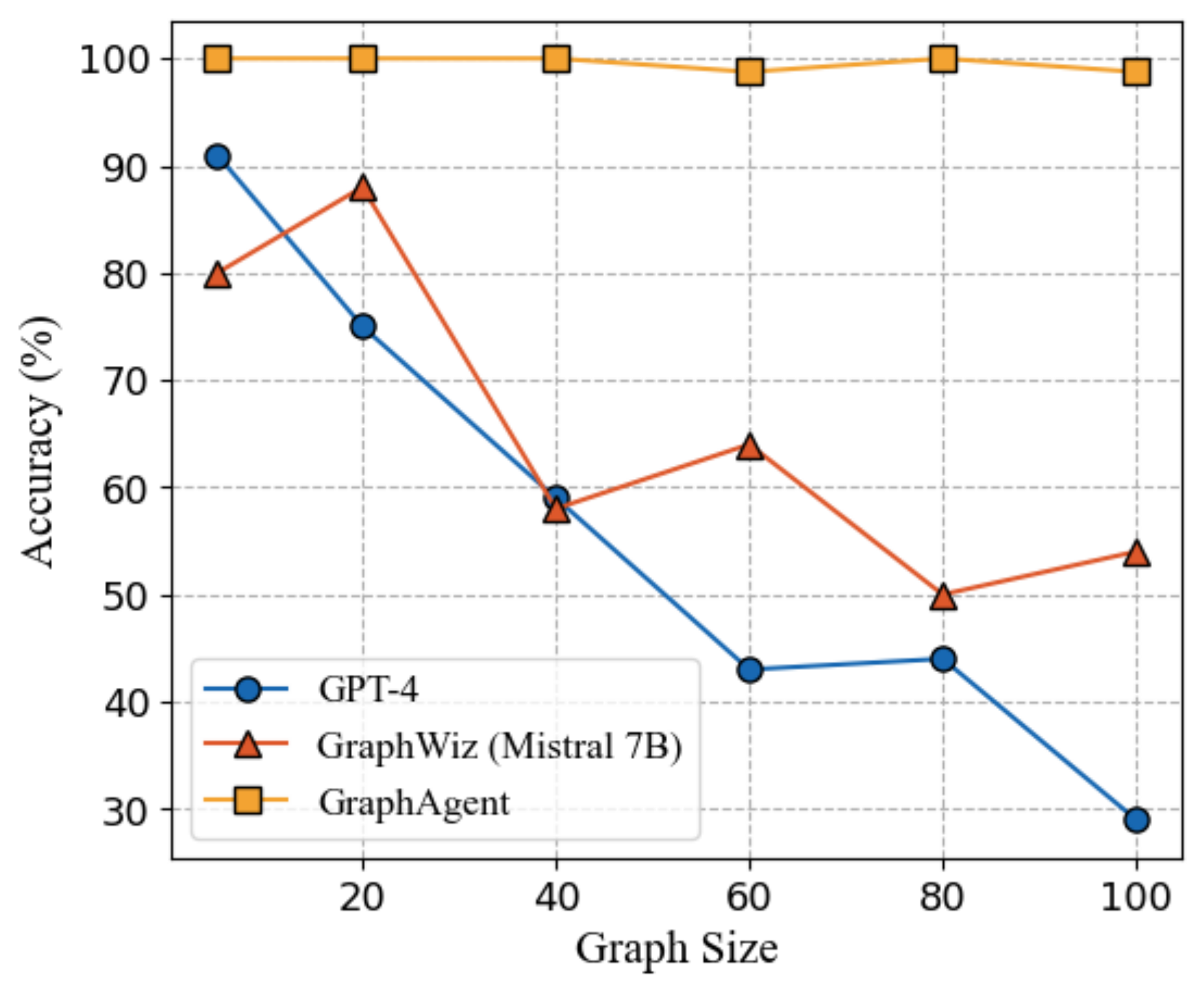}
        \label{fig:cycle}
    }
    \hspace{2pt} 
    \subfigure[Shortest Path]{
        \includegraphics[width=0.21\textwidth]{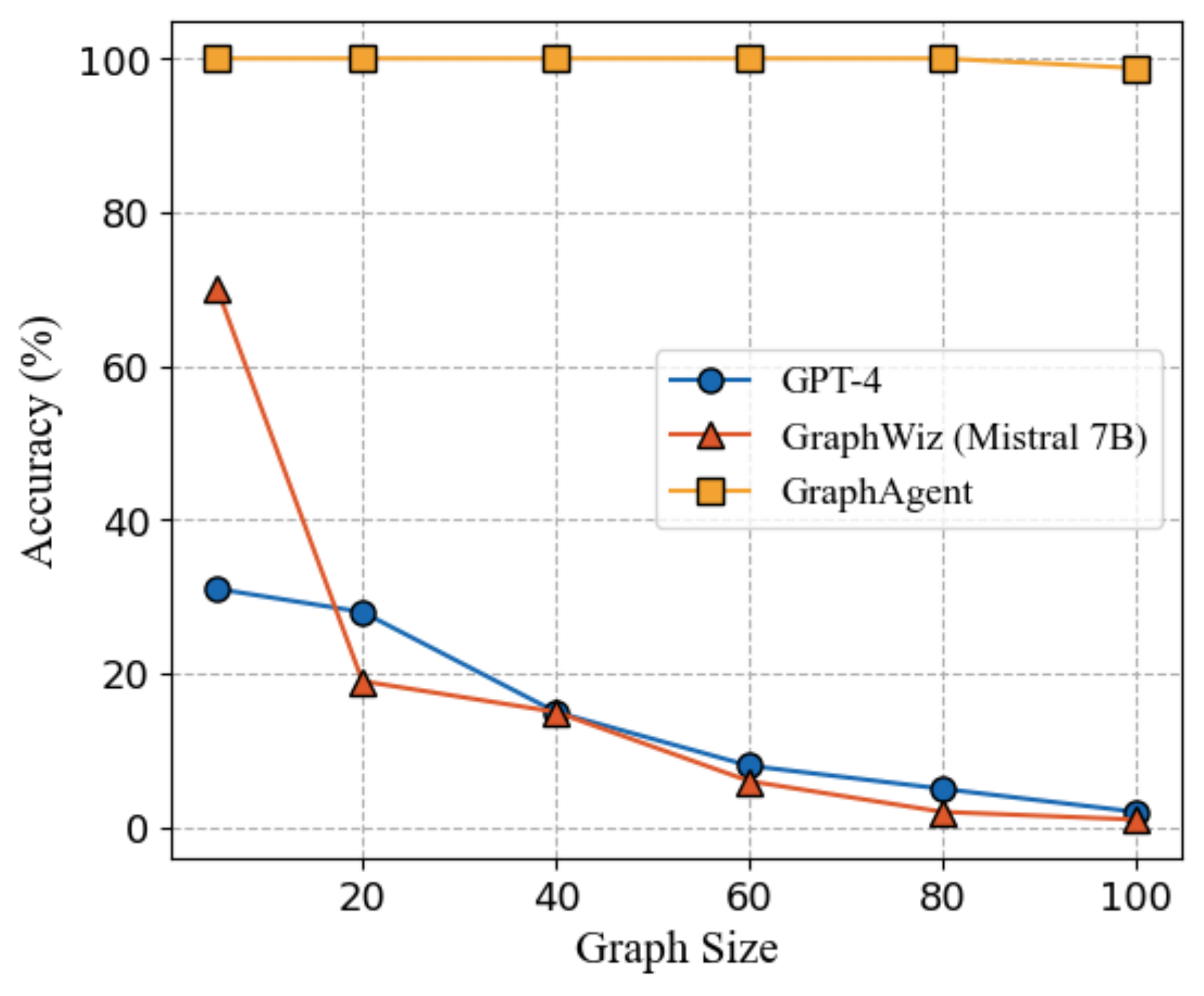}
        \label{fig:shortest}
    }
    \caption{Performance of GraphAgent-Reasoner, GPT4(2 shot) and GraphWiz(Mistral 7B) on cycle detection and shortest path problems with different graph sizes.}
    \label{fig:comparison}
    \vspace{-1em}
\end{figure}

As the number of nodes increases, the graph structures become more complex, making the solution of graph problems increasingly difficult. To investigate how the performance of models varies with increasing problem complexity, we conduct experiments on cycle detection and shortest path problems, gradually increasing the number of nodes from 5 to 100. The results are presented in Figure~\ref{fig:comparison}. 

We see with the number of nodes increasing, both ChatGPT-4 and Graphwiz exhibit a significant decline in performance. However, the accuracy of GAR remains stable, almost unaffected by the graph size, demonstrating robust scalability. Although the scale of the graph is increasing, the information processed by each agent has not significantly increased. Each agent still only handles its own information and communicates with neighboring agents. We observe that GAR occasionally makes errors in specific cases, likely due to the increasing communication rounds as the number of nodes and edges grows. Even when handling simple node-centric tasks, a single agent still has the potential to make mistakes. 
Therefore, as the number of agents and communication rounds increases, the overall likelihood of errors also rises. This can be improved by enhancing the capability of individual agents (such as using stronger LLMs as the underlying reasoning model) or by more finely designed prompts.

\subsection{Performance on Large-Scale 
Graphs}
In this experiment, we evaluate the performance of current LLMs on large-scale graphs. The largest graph size handled by existing graph reasoning work is 100 nodes~\citep{chen2024graphwiz}, which is still far from sufficient for real-world graph reasoning scenarios. To evaluate the reasoning performance of existing models on larger graphs, we conduct shortest path experiments on graphs with 100, 200, 500, and 1000 nodes. Due to the excessively long input text (reaching 16,000 tokens for 1000 nodes) and the money cost, we only create 20 test samples for each graph size. The results are shown in Table~\ref{tab:large-scale}. 

\begin{table}[h!]
    \centering
    \caption{Performance on large-scale graphs dealing with shortest path problems. x/20 indicates that out of 20 test samples, x samples are correct. NA signifies that testing could not be conducted due to the fact that the context length limit is exceeded.}
    \resizebox{\linewidth}{!}{%
    \begin{tabular}{lrrrr}
    \toprule
    Graph Size & 100 & 200 & 500 & 1000 \\ 
    \midrule
    Graphwiz (LLaMA 2-7B) & 0/20 & 0/20 & NA & NA\\
    Graphwiz (LLaMA 2-7B-DPO) & 0/20 & 0/20 & NA & NA\\
    Chatgpt-3.5-turbo-16k & 0/20 & 0/20 & 0/20 & 0/20\\ 
    Chatgpt-4-32k & 0/20 & 1/20 & 0/20 & 0/20\\ 
    GraphAgent-Reasoner & \textbf{20/20} & \textbf{20/20} & \textbf{20/20} & \textbf{18/20} \\
    \bottomrule
    \end{tabular}
    }
    \label{tab:large-scale}
    \vspace{-0.5em}
    \end{table}

We see the two GraphWiz models fine-tuned on the LLaMA2-7B~\citep{Touvron2023llama2} base model are unable to handle graphs with 500 or more nodes due to the context length limitation (the context length limit for Llama2 is 4096 tokens). Although ChatGPT-3.5-turbo-16k and ChatGPT-4-32k can manage longer contexts, they output wrong answers in almost all test samples, with only ChatGPT-4-32k being correct in one 200 nodes test sample. In contrast, GAR maintains a high accuracy in large-scale graph, only failed in two 1000-node test samples, further demonstrating its robust scalability.

\subsection{Case Study}
\label{case}
\begin{figure}[htbp]
\centering
\includegraphics[width=0.5\textwidth]{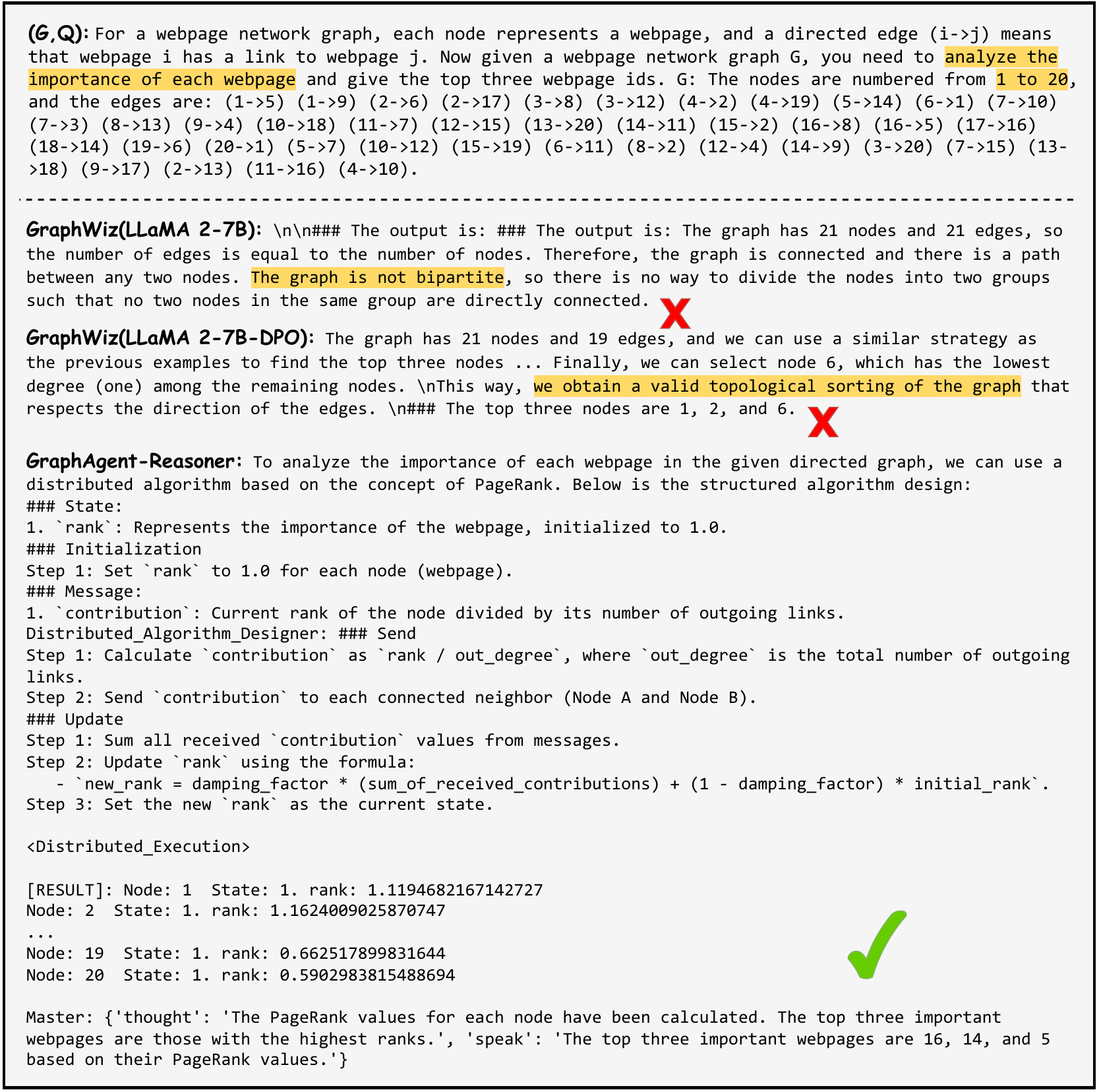}
\caption{The importance analysis in webpage network. While the GraphWiz fails due to incorrect graph assessments, GAR correctly uses the PageRank algorithm to identify nodes 16, 14, and 5 as the most important.}
\label{fig:case1}
\end{figure}

In this experiment, we explore the application of two graph reasoning models, Graphwiz and GAR, in real-world graph reasoning scenarios. We present a case study of webpage importance analysis in Figure~\ref{fig:case1}.

Although GraphWiz performed well on fine-tuned tasks, it exhibits severe overfitting when faced with real-world graph problems, failing to apply the graph reasoning knowledge learned during the fine-tuning phase. 
Since GraphWiz uses a consistent graph node description, the sentence "The nodes are numbered from 0 to ..." appears across all datasets during the multi-task instruction tuning. 
When the actual problem has nodes numbered from 1 to 20, it still assumes the existence of node 0. Both GraphWiz models output that the graph has 21 nodes with an incorrect number of edges. 

Furthermore, neither of the two GraphWiz models recognizes that this is a problem associated with web page importance ranking. 
Instead, they approach it as the bipartite graph check or topological sort problems they had been fine-tuned on. Additionally, neither model generates an explicit and correct reasoning path. 
These observations indicate that there is still a significant gap between excelling in classic graph reasoning tasks and effectively solving real-world graph reasoning problems. In contrast, GAR correctly identifies that the problem should be solved using knowledge related to PageRank~\citep{yang2024pagerank} and designs an algorithm that adhered to the distributed paradigm (Note: the distributed algorithm library does not contain a PageRank algorithm template). 
GAR then assigns the algorithm to agent nodes for execution, ultimately obtaining the PageRank value for each node and arriving at the correct conclusion. 
Through the distributed paradigm, GAR effectively bridges the powerful knowledge learned by LLMs with the solving of real-world graph reasoning problems, which enables it to flexibly handle practical issues in a distributed manner. 
This case study demonstrates the feasibility of using GAR to solve real-world graph reasoning problems, indicating its substantial practical applicability and offering researchers and practitioners a powerful framework to address such tasks.


\section{Conclusion}

We first summarize three key issues faced by existing LLMs in graph reasoning tasks: limited graph scale, poor performance, and the lack of explicit reasoning paths. We then reflect on the limitations of a single LLM in addressing graph reasoning problems, such as the graph structures being too complex to memorize and understand and the overwhelming information in real-world graph reasoning scenarios. To address these challenges, we propose GraphAgent-Reasoner, a framework based on multi-agent collaboration to solve graph reasoning problems. This framework demonstrates superior accuracy and scalability, significantly surpassing existing closed-source and fine-tuned open-source models. Our experiments show its robust scalability, maintaining high accuracy on large graphs (up to 1,000 nodes). Our case study on webpage importance analysis further illustrates its capability to handle real-world graph reasoning problems. Future work will focus on designing more accurate and scalable LLM-based multi-agent graph reasoning frameworks, aiming to apply them to larger and more complex real-world reasoning scenarios.

\section{Acknowledgments}
This work was supported in part  by Ant Group through Ant Research Intern Program, by National Natural Science Foundation of China (No. 92470128, No. U2241212), by National Science and Technology Major Project (2022ZD0114802), by Beijing Outstanding Young Scientist Program No.BJJWZYJH012019100020098, by the National Key Research and Development Plan of China (2023YFB4502305). 
We also wish to acknowledge the support provided by the fund for building world-class universities (disciplines) of Renmin University of China, by Engineering Research Center of Next-Generation Intelligent Search and Recommendation, Ministry of Education, by Intelligent Social Governance Interdisciplinary Platform, Major Innovation \& Planning Interdisciplinary Platform for the “Double-First Class” Initiative, Public Policy and Decision-making Research Lab, and Public Computing Cloud, Renmin University of China.
The work was partially done at Gaoling School of Artificial Intelligence, Beijing Key Laboratory of Big Data Management and Analysis Methods, MOE Key Lab of Data Engineering and Knowledge Engineering, and Pazhou Laboratory (Huangpu), Guangzhou, Guangdong 510555, China.

\bibliography{aaai2026}

@inproceedings{brown2020language,
  author       = {Brown, Tom and Mann, Benjamin and Ryder, Nick and Subbiah, Melanie and Kaplan, Jared D and Dhariwal, Prafulla and Neelakantan, Arvind and Shyam, Pranav and Sastry, Girish and Askell, Amanda and others},
  editor       = {Hugo Larochelle and
                  Marc'Aurelio Ranzato and
                  Raia Hadsell and
                  Maria{-}Florina Balcan and
                  Hsuan{-}Tien Lin},
  title        = {Language Models are Few-Shot Learners},
  booktitle    = {Advances in Neural Information Processing Systems 33: Annual Conference
                  on Neural Information Processing Systems 2020, NeurIPS 2020, December
                  6-12, 2020, virtual},
  year         = {2020},
  url          = {https://proceedings.neurips.cc/paper/2020/hash/1457c0d6bfcb4967418bfb8ac142f64a-Abstract.html}
}

@article{chen2024exploring,
  author       = {Zhikai Chen and
                  Haitao Mao and
                  Hang Li and
                  Wei Jin and
                  Hongzhi Wen and
                  Xiaochi Wei and
                  Shuaiqiang Wang and
                  Dawei Yin and
                  Wenqi Fan and
                  Hui Liu and
                  Jiliang Tang},
  title        = {Exploring the Potential of Large Language Models (LLMs)in Learning
                  on Graphs},
  journal      = {{SIGKDD} Explor.},
  volume       = {25},
  number       = {2},
  pages        = {42--61},
  year         = {2023},
  url          = {https://doi.org/10.1145/3655103.3655110}
}

@inproceedings{wang2024can,
  author       = {Heng Wang and
                  Shangbin Feng and
                  Tianxing He and
                  Zhaoxuan Tan and
                  Xiaochuang Han and
                  Yulia Tsvetkov},
  editor       = {Alice Oh and
                  Tristan Naumann and
                  Amir Globerson and
                  Kate Saenko and
                  Moritz Hardt and
                  Sergey Levine},
  title        = {Can Language Models Solve Graph Problems in Natural Language?},
  booktitle    = {Advances in Neural Information Processing Systems 36: Annual Conference
                  on Neural Information Processing Systems 2023, NeurIPS 2023, New Orleans,
                  LA, USA, December 10 - 16, 2023},
  year         = {2023},
  url          = {http://papers.nips.cc/paper\_files/paper/2023/hash/622afc4edf2824a1b6aaf5afe153fa93-Abstract-Conference.html}
}

@article{Liu2023LostIT,
  title={Lost in the Middle: How Language Models Use Long Contexts},
  author={Nelson F. Liu and Kevin Lin and John Hewitt and Ashwin Paranjape and Michele Bevilacqua and Fabio Petroni and Percy Liang},
  journal={Transactions of the Association for Computational Linguistics},
  year={2023},
  volume={12},
  pages={157-173},
  url={https://api.semanticscholar.org/CorpusID:259360665}
}

@article{achiam2023gpt,
  author       = {OpenAI},
  title        = {{GPT-4} Technical Report},
  journal      = {CoRR},
  volume       = {abs/2303.08774},
  year         = {2023},
  url          = {https://doi.org/10.48550/arXiv.2303.08774}
}

@article{stokes2020deep,
  title={A deep learning approach to antibiotic discovery},
  author={Stokes, Jonathan M and Yang, Kevin and Swanson, Kyle and Jin, Wengong and Cubillos-Ruiz, Andres and Donghia, Nina M and MacNair, Craig R and French, Shawn and Carfrae, Lindsey A and Bloom-Ackermann, Zohar and others},
  journal={Cell},
  volume={180},
  number={4},
  pages={688--702},
  year={2020},
  publisher={Elsevier}
}

@article{jiang2022graph,
  author       = {Weiwei Jiang and
                  Jiayun Luo},
  title        = {Graph neural network for traffic forecasting: {A} survey},
  journal      = {Expert Syst. Appl.},
  volume       = {207},
  pages        = {117921},
  year         = {2022},
  url          = {https://doi.org/10.1016/j.eswa.2022.117921}
}

@article{motie2023financial,
  author       = {Soroor Motie and
                  Bijan Raahemi},
  title        = {Financial fraud detection using graph neural networks: {A} systematic
                  review},
  journal      = {Expert Syst. Appl.},
  volume       = {240},
  pages        = {122156},
  year         = {2024},
  url          = {https://doi.org/10.1016/j.eswa.2023.122156}
}

@inproceedings{fatemi2023talk,
  author       = {Bahare Fatemi and
                  Jonathan Halcrow and
                  Bryan Perozzi},
  title        = {Talk like a Graph: Encoding Graphs for Large Language Models},
  booktitle    = {The Twelfth International Conference on Learning Representations,
                  {ICLR} 2024, Vienna, Austria, May 7-11, 2024},
  publisher    = {OpenReview.net},
  year         = {2024},
  url          = {https://openreview.net/forum?id=IuXR1CCrSi}
}

@article{perozzi2024let,
  author       = {Bryan Perozzi and
                  Bahare Fatemi and
                  Dustin Zelle and
                  Anton Tsitsulin and
                  Seyed Mehran Kazemi and
                  Rami Al{-}Rfou and
                  Jonathan Halcrow},
  title        = {Let Your Graph Do the Talking: Encoding Structured Data for LLMs},
  journal      = {CoRR},
  volume       = {abs/2402.05862},
  year         = {2024},
  url          = {https://doi.org/10.48550/arXiv.2402.05862}
}

@inproceedings{chen2024graphwiz,
  author       = {Nuo Chen and
                  Yuhan Li and
                  Jianheng Tang and
                  Jia Li},
  editor       = {Ricardo Baeza{-}Yates and
                  Francesco Bonchi},
  title        = {GraphWiz: An Instruction-Following Language Model for Graph Computational
                  Problems},
  booktitle    = {Proceedings of the 30th {ACM} {SIGKDD} Conference on Knowledge Discovery
                  and Data Mining, {KDD} 2024, Barcelona, Spain, August 25-29, 2024},
  pages        = {353--364},
  publisher    = {{ACM}},
  year         = {2024},
  url          = {https://doi.org/10.1145/3637528.3672010}
}

@article{agentscope,
  author       = {Dawei Gao and
                  Zitao Li and
                  Weirui Kuang and
                  Xuchen Pan and
                  Daoyuan Chen and
                  Zhijian Ma and
                  Bingchen Qian and
                  Liuyi Yao and
                  Lin Zhu and
                  Chen Cheng and
                  Hongzhu Shi and
                  Yaliang Li and
                  Bolin Ding and
                  Jingren Zhou},
  title        = {AgentScope: {A} Flexible yet Robust Multi-Agent Platform},
  journal      = {CoRR},
  volume       = {abs/2402.14034},
  year         = {2024},
  url          = {https://doi.org/10.48550/arXiv.2402.14034}
}

@inproceedings{yao2023react,
  author       = {Shunyu Yao and
                  Jeffrey Zhao and
                  Dian Yu and
                  Nan Du and
                  Izhak Shafran and
                  Karthik R. Narasimhan and
                  Yuan Cao},
  title        = {ReAct: Synergizing Reasoning and Acting in Language Models},
  booktitle    = {The Eleventh International Conference on Learning Representations,
                  {ICLR} 2023, Kigali, Rwanda, May 1-5, 2023},
  publisher    = {OpenReview.net},
  year         = {2023},
  url          = {https://openreview.net/forum?id=WE\_vluYUL-X}
}

@article{huang2024can,
  author       = {Jin Huang and
                  Xingjian Zhang and
                  Qiaozhu Mei and
                  Jiaqi Ma},
  title        = {Can LLMs Effectively Leverage Graph Structural Information through
                  Prompts, and Why?},
  journal      = {Trans. Mach. Learn. Res.},
  volume       = {2024},
  year         = {2024},
  url          = {https://openreview.net/forum?id=L2jRavXRxs}
}

@article{guo2023gpt4graph,
  author       = {Jiayan Guo and
                  Lun Du and
                  Hengyu Liu},
  title        = {GPT4Graph: Can Large Language Models Understand Graph Structured Data
                  ? An Empirical Evaluation and Benchmarking},
  journal      = {CoRR},
  volume       = {abs/2305.15066},
  year         = {2023},
  url          = {https://doi.org/10.48550/arXiv.2305.15066}
}

@inproceedings{vaswani2017attention,
  author       = {Ashish Vaswani and
                  Noam Shazeer and
                  Niki Parmar and
                  Jakob Uszkoreit and
                  Llion Jones and
                  Aidan N. Gomez and
                  Lukasz Kaiser and
                  Illia Polosukhin},
  editor       = {Isabelle Guyon and
                  Ulrike von Luxburg and
                  Samy Bengio and
                  Hanna M. Wallach and
                  Rob Fergus and
                  S. V. N. Vishwanathan and
                  Roman Garnett},
  title        = {Attention is All you Need},
  booktitle    = {Advances in Neural Information Processing Systems 30: Annual Conference
                  on Neural Information Processing Systems 2017, December 4-9, 2017,
                  Long Beach, CA, {USA}},
  pages        = {5998--6008},
  year         = {2017},
  url          = {https://proceedings.neurips.cc/paper/2017/hash/3f5ee243547dee91fbd053c1c4a845aa-Abstract.html}
}

@article{meng2024survey,
  author       = {Lingkai Meng and
                  Yu Shao and
                  Long Yuan and
                  Longbin Lai and
                  Peng Cheng and
                  Xue Li and
                  Wenyuan Yu and
                  Wenjie Zhang and
                  Xuemin Lin and
                  Jingren Zhou},
  title        = {A Survey of Distributed Graph Algorithms on Massive Graphs},
  journal      = {CoRR},
  volume       = {abs/2404.06037},
  year         = {2024},
  url          = {https://doi.org/10.48550/arXiv.2404.06037}
}

@inproceedings{zheng2024executing,
  author       = {Xin Zheng and
                  Qiming Zhu and
                  Hongyu Lin and
                  Yaojie Lu and
                  Xianpei Han and
                  Le Sun},
  editor       = {Nicoletta Calzolari and
                  Min{-}Yen Kan and
                  V{\'{e}}ronique Hoste and
                  Alessandro Lenci and
                  Sakriani Sakti and
                  Nianwen Xue},
  title        = {Executing Natural Language-Described Algorithms with Large Language
                  Models: An Investigation},
  booktitle    = {Proceedings of the 2024 Joint International Conference on Computational
                  Linguistics, Language Resources and Evaluation, {LREC/COLING} 2024,
                  20-25 May, 2024, Torino, Italy},
  pages        = {6752--6837},
  publisher    = {{ELRA} and {ICCL}},
  year         = {2024},
  url          = {https://aclanthology.org/2024.lrec-main.596}
}

@article{jojic2024gpt,
  author       = {Lingkai Meng and
                  Yu Shao and
                  Long Yuan and
                  Longbin Lai and
                  Peng Cheng and
                  Xue Li and
                  Wenyuan Yu and
                  Wenjie Zhang and
                  Xuemin Lin and
                  Jingren Zhou},
  title        = {A Survey of Distributed Graph Algorithms on Massive Graphs},
  journal      = {CoRR},
  volume       = {abs/2404.06037},
  year         = {2024},
  url          = {https://doi.org/10.48550/arXiv.2404.06037}
}

@inproceedings{hong2024metagpt,
  author       = {Sirui Hong and
                  Mingchen Zhuge and
                  Jonathan Chen and
                  Xiawu Zheng and
                  Yuheng Cheng and
                  Jinlin Wang and
                  Ceyao Zhang and
                  Zili Wang and
                  Steven Ka Shing Yau and
                  Zijuan Lin and
                  Liyang Zhou and
                  Chenyu Ran and
                  Lingfeng Xiao and
                  Chenglin Wu and
                  J{\"{u}}rgen Schmidhuber},
  title        = {MetaGPT: Meta Programming for {A} Multi-Agent Collaborative Framework},
  booktitle    = {The Twelfth International Conference on Learning Representations,
                  {ICLR} 2024, Vienna, Austria, May 7-11, 2024},
  publisher    = {OpenReview.net},
  year         = {2024},
  url          = {https://openreview.net/forum?id=VtmBAGCN7o}
}

@inproceedings{qian2024chatdev,
  author       = {Chen Qian and
                  Wei Liu and
                  Hongzhang Liu and
                  Nuo Chen and
                  Yufan Dang and
                  Jiahao Li and
                  Cheng Yang and
                  Weize Chen and
                  Yusheng Su and
                  Xin Cong and
                  Juyuan Xu and
                  Dahai Li and
                  Zhiyuan Liu and
                  Maosong Sun},
  editor       = {Lun{-}Wei Ku and
                  Andre Martins and
                  Vivek Srikumar},
  title        = {ChatDev: Communicative Agents for Software Development},
  booktitle    = {Proceedings of the 62nd Annual Meeting of the Association for Computational
                  Linguistics (Volume 1: Long Papers), {ACL} 2024, Bangkok, Thailand,
                  August 11-16, 2024},
  pages        = {15174--15186},
  publisher    = {Association for Computational Linguistics},
  year         = {2024},
  url          = {https://doi.org/10.18653/v1/2024.acl-long.810}
}

@article{dong2023self,
  author       = {Yihong Dong and
                  Xue Jiang and
                  Zhi Jin and
                  Ge Li},
  title        = {Self-collaboration Code Generation via ChatGPT},
  journal      = {CoRR},
  volume       = {abs/2304.07590},
  year         = {2023},
  url          = {https://doi.org/10.48550/arXiv.2304.07590}
}

@inproceedings{zhang2024build,
  author       = {Hongxin Zhang and
                  Weihua Du and
                  Jiaming Shan and
                  Qinhong Zhou and
                  Yilun Du and
                  Joshua B. Tenenbaum and
                  Tianmin Shu and
                  Chuang Gan},
  title        = {Building Cooperative Embodied Agents Modularly with Large Language
                  Models},
  booktitle    = {The Twelfth International Conference on Learning Representations,
                  {ICLR} 2024, Vienna, Austria, May 7-11, 2024},
  publisher    = {OpenReview.net},
  year         = {2024},
  url          = {https://openreview.net/forum?id=EnXJfQqy0K}
}

@inproceedings{zhao2024roco,
  author       = {Zhao Mandi and
                  Shreeya Jain and
                  Shuran Song},
  title        = {RoCo: Dialectic Multi-Robot Collaboration with Large Language Models},
  booktitle    = {{IEEE} International Conference on Robotics and Automation, {ICRA}
                  2024, Yokohama, Japan, May 13-17, 2024},
  pages        = {286--299},
  publisher    = {{IEEE}},
  year         = {2024},
  url          = {https://doi.org/10.1109/ICRA57147.2024.10610855}
}

@inproceedings{chen2024scalable,
  author       = {Yongchao Chen and
                  Jacob Arkin and
                  Yang Zhang and
                  Nicholas Roy and
                  Chuchu Fan},
  title        = {Scalable Multi-Robot Collaboration with Large Language Models: Centralized
                  or Decentralized Systems?},
  booktitle    = {{IEEE} International Conference on Robotics and Automation, {ICRA}
                  2024, Yokohama, Japan, May 13-17, 2024},
  pages        = {4311--4317},
  publisher    = {{IEEE}},
  year         = {2024},
  url          = {https://doi.org/10.1109/ICRA57147.2024.10610676}
}

@inproceedings{xiong2023examing,
  author       = {Kai Xiong and
                  Xiao Ding and
                  Yixin Cao and
                  Ting Liu and
                  Bing Qin},
  editor       = {Houda Bouamor and
                  Juan Pino and
                  Kalika Bali},
  title        = {Examining Inter-Consistency of Large Language Models Collaboration:
                  An In-depth Analysis via Debate},
  booktitle    = {Findings of the Association for Computational Linguistics: {EMNLP}
                  2023, Singapore, December 6-10, 2023},
  pages        = {7572--7590},
  publisher    = {Association for Computational Linguistics},
  year         = {2023},
  url          = {https://doi.org/10.18653/v1/2023.findings-emnlp.508}
}

@inproceedings{chan2024chateval,
  author       = {Chi{-}Min Chan and
                  Weize Chen and
                  Yusheng Su and
                  Jianxuan Yu and
                  Wei Xue and
                  Shanghang Zhang and
                  Jie Fu and
                  Zhiyuan Liu},
  title        = {ChatEval: Towards Better LLM-based Evaluators through Multi-Agent
                  Debate},
  booktitle    = {The Twelfth International Conference on Learning Representations,
                  {ICLR} 2024, Vienna, Austria, May 7-11, 2024},
  publisher    = {OpenReview.net},
  year         = {2024},
  url          = {https://openreview.net/forum?id=FQepisCUWu}
}

@article{huang2023hallucination,
  author       = {Lei Huang and
                  Weijiang Yu and
                  Weitao Ma and
                  Weihong Zhong and
                  Zhangyin Feng and
                  Haotian Wang and
                  Qianglong Chen and
                  Weihua Peng and
                  Xiaocheng Feng and
                  Bing Qin and
                  Ting Liu},
  title        = {A Survey on Hallucination in Large Language Models: Principles, Taxonomy,
                  Challenges, and Open Questions},
  journal      = {CoRR},
  volume       = {abs/2311.05232},
  year         = {2023},
  url          = {https://doi.org/10.48550/arXiv.2311.05232}
}

@article{guo2024multiagent,
  author       = {Taicheng Guo and
                  Xiuying Chen and
                  Yaqi Wang and
                  Ruidi Chang and
                  Shichao Pei and
                  Nitesh V. Chawla and
                  Olaf Wiest and
                  Xiangliang Zhang},
  title        = {Large Language Model based Multi-Agents: {A} Survey of Progress and
                  Challenges},
  journal      = {CoRR},
  volume       = {abs/2402.01680},
  year         = {2024},
  url          = {https://doi.org/10.48550/arXiv.2402.01680}
}

@article{meadows2023math,
  author       = {Jordan Meadows and
                  Marco Valentino and
                  Andr{\'{e}} Freitas},
  title        = {Generating Mathematical Derivations with Large Language Models},
  journal      = {CoRR},
  volume       = {abs/2307.09998},
  year         = {2023},
  url          = {https://doi.org/10.48550/arXiv.2307.09998}
}

@inproceedings{Madaan2022commonsense,
  author       = {Aman Madaan and
                  Shuyan Zhou and
                  Uri Alon and
                  Yiming Yang and
                  Graham Neubig},
  editor       = {Yoav Goldberg and
                  Zornitsa Kozareva and
                  Yue Zhang},
  title        = {Language Models of Code are Few-Shot Commonsense Learners},
  booktitle    = {Proceedings of the 2022 Conference on Empirical Methods in Natural
                  Language Processing, {EMNLP} 2022, Abu Dhabi, United Arab Emirates,
                  December 7-11, 2022},
  pages        = {1384--1403},
  publisher    = {Association for Computational Linguistics},
  year         = {2022},
  url          = {https://doi.org/10.18653/v1/2022.emnlp-main.90}
}

@inproceedings{Creswell2023selection,
  author       = {Antonia Creswell and
                  Murray Shanahan and
                  Irina Higgins},
  title        = {Selection-Inference: Exploiting Large Language Models for Interpretable
                  Logical Reasoning},
  booktitle    = {The Eleventh International Conference on Learning Representations,
                  {ICLR} 2023, Kigali, Rwanda, May 1-5, 2023},
  publisher    = {OpenReview.net},
  year         = {2023},
  url          = {https://openreview.net/forum?id=3Pf3Wg6o-A4}
}

@article{chai2023graphllm,
  author       = {Ziwei Chai and
                  Tianjie Zhang and
                  Liang Wu and
                  Kaiqiao Han and
                  Xiaohai Hu and
                  Xuanwen Huang and
                  Yang Yang},
  title        = {GraphLLM: Boosting Graph Reasoning Ability of Large Language Model},
  journal      = {CoRR},
  volume       = {abs/2310.05845},
  year         = {2023},
  url          = {https://doi.org/10.48550/arXiv.2310.05845}
}

@inproceedings{Wei2024GITAGT,
  title={GITA: Graph to Visual and Textual Integration for Vision-Language Graph Reasoning},
  author={Yanbin Wei and Shuai Fu and Weisen Jiang and James T. Kwok and Yu Zhang},
  year={2024},
  url={https://api.semanticscholar.org/CorpusID:267413180}
}

@article{Touvron2023llama2,
  author       = {Hugo Touvron and
                  Louis Martin and
                  Kevin Stone and
                  Peter Albert and others},
  title        = {Llama 2: Open Foundation and Fine-Tuned Chat Models},
  journal      = {CoRR},
  volume       = {abs/2307.09288},
  year         = {2023},
  url          = {https://doi.org/10.48550/arXiv.2307.09288}
}

@article{yang2024pagerank,
  author       = {Mingji Yang and
                  Hanzhi Wang and
                  Zhewei Wei and
                  Sibo Wang and
                  Ji{-}Rong Wen},
  title        = {Efficient Algorithms for Personalized PageRank Computation: {A} Survey},
  journal      = {{IEEE} Trans. Knowl. Data Eng.},
  volume       = {36},
  number       = {9},
  pages        = {4582--4602},
  year         = {2024},
  url          = {https://doi.org/10.1109/TKDE.2024.3376000}
}

\clearpage

\onecolumn{\section{Appendix}
\subsection{The GraphInstruct Dataset}
\label{app:dataset}
The statistics and detailed information of GraphInstruct are shown in Table~\ref{tab:graphinstruct}. Hamilton Path and Subgraph Matching are NP-complete problems. However, there are no well-established distributed algorithms for these two problems, and GraphAgent-Reasoner struggles to solve them. The following shows its attempt on the Hamilton Path problem.

\begin{table*}[htp!]
\centering
\caption{The detailed information of GraphInstruct dataset.}
\resizebox{0.9\linewidth}{!}{%
\begin{tabular}{p{0.25\textwidth}p{0.5\textwidth}cc}
\toprule
\textbf{Problem}  &  \textbf{Definition} & \textbf{Node Range} & \textbf{Test Size}\\ 
\midrule
Cycle Detection &  Detect if a given graph $\mathcal{G}$ contains any cycles. &  [2, 100]  & 400 \\ \midrule
Connectivity &  Assess if two nodes $u$ and $v$ in a given graph $\mathcal{G}$ are connected via a path. & [2, 100] & 400 \\ \midrule

Bipartite Graph Check & Judge if a given graph $\mathcal{G}$ is bipartite. &  [2, 100]  & 400 \\ \midrule

Topological Sort & Find a topological ordering of vertices in a directed acyclic graph $\mathcal{G}$. & [2, 50] & 400 \\ \midrule

Shortest Path & Compute the shortest path between two specific nodes $u$ and $v$ in a given graph $\mathcal{G}$.  & [2, 100] & 400 \\ \midrule

Maximum Triangle Sum & Find the maximum sum of weights for any connected triplet of vertices in a given graph $\mathcal{G}$.  & [2, 25] & 400 \\ \midrule

Maximum Flow & Calculate the maximum flow from a source node $s$ to a sink node $t$ in a directed graph $\mathcal{G}$. & [2, 50] & 400 \\ \midrule

Hamilton Path &  Determine if a given graph $\mathcal{G}$ has a Hamiltonian path that visits each vertex exactly once. & [2, 50] & 400 \\ \midrule

Subgraph Matching & Verify if there exists a subgraph in $\mathcal{G}$ that is isomorphic to a given graph $\mathcal{G}'$. & [2, 30] & 400 \\
\bottomrule
\end{tabular}
}
\label{tab:graphinstruct}
\vspace{-0.5em}
\end{table*}

\textbf{Hamilton Path Execution Example.}

Problem Description: 

Determine whether or not there is a Hamiltonian path in an undirected graph. In an undirected graph, (i,j) means that node i and node j are connected with an undirected edge. Given a graph, you need to output Yes or No, indicating whether there is a Hamiltonian path in the graph. Q: The nodes are numbered from 0 to 5, and the edges are: (0, 3) (0, 1) (0, 2) (0, 4) (1, 5) (1, 4) (1, 2) (1, 3) (2, 4) (2, 5) (3, 5) (3, 4). Is there a Hamiltonian path in this graph?

Execution Process:

\begin{tcolorbox}[colback=gray!10, colframe=black, rounded corners, boxrule=1.5pt, listing only, listing options={basicstyle=\small\ttfamily, breaklines=true, columns=fullflexible, numbers=none, xleftmargin=0pt}, left=2mm, right=2mm, top=1mm, bottom=1mm]
\begin{lstlisting}[numbers=none, xleftmargin=0pt]
### State
1. `visited`: Boolean indicating if the node has been visited.
2. `path_length`: Integer representing the current length of the path.
3. `max_path_length`: Integer tracking the longest path found.
 ### Initialization
Step 1: Set `visited = False`, `path_length = 0`, and `max_path_length = 1` for all nodes.  
Step 2: For the initial node (e.g., Node Id: 0), set `visited = True` and
`path_length = 1`.
Distributed_Algorithm_Designer: ### Message
1. `path_length`: The current path length from the sender.
2. `max_path_length`: The maximum path length known to the sender.
3. `visited_flag`: Boolean indicating whether the sender has visited the node.
### Send
Step 1: For each neighbor (Node A, Node B), if `visited` is `False`, construct the message as follows:  
- Message to Node A: `(Node Id: <node_id>, path_length + 1, max(max_path_length, path_length + 1))`.  
- Message to Node B: `(Node Id: <node_id>, path_length + 1, max(max_path_length, path_length + 1))`.  
Step 2: Send the constructed messages to both neighbors.
\end{lstlisting}
\end{tcolorbox}

\begin{tcolorbox}[colback=gray!10, colframe=black, rounded corners, boxrule=1.5pt, listing only, listing options={basicstyle=\small\ttfamily, breaklines=true, columns=fullflexible, numbers=none, xleftmargin=0pt}, left=2mm, right=2mm, top=1mm, bottom=1mm]
\begin{lstlisting}[numbers=none, xleftmargin=0pt]
### Example
## Input
Node Id: 0  
State:  
1. visited: False  
2. path_length: 2  
3. max_path_length: 3  
Neighbor Information:  
Connected to:  
Node A  
Node B  
## Process
Step 1: Check neighbors Node A and Node B. Since `visited` is 
`False`, proceed to construct messages.  
- For Node A:  
  - Calculate new `path_length`: `2 + 1 = 3`  
  - Calculate `max_path_length`: `max(3, 2 + 1) = 3`  
- For Node B:  
  - Same calculations apply as for Node A.  
Step 2: Construct messages:  
- Message to Node A: `(Node Id: 0, 3, 3)`  
- Message to Node B: `(Node Id: 0, 3, 3)`  
## Output
Message sent to Node A:  
1. path_length: 3  
2. max_path_length: 3  
3. visited_flag: False  
Message sent to Node B:  
1. path_length: 3  
2. max_path_length: 3  
3. visited_flag: False  
### Update
Step 1: For each received message, check the `visited_flag`. If it is 
`False`, proceed to the next steps.  
Step 2: Update `visited` to `True` for the current node.  
Step 3: Update `path_length` to the maximum of its current value and
the received `path_length` from the message, incrementing by 1 if the
message indicates a potential continuation of the path.  
Step 4: Update `max_path_length` to the maximum of its current value 
and the received `max_path_length` from the message.  
Step 5: If any messages were processed, trigger a re-evaluation of 
outgoing messages to neighbors, if applicable.
### Example
## Input
Node Id: 1  
State:  
1. visited: False  
2. path_length: 1  
3. max_path_length: 2  
Received Messages:  
Message 1:  
1. path_length: 2  
2. max_path_length: 3  
3. visited_flag: False  
Message 2:  
1. path_length: 3  
2. max_path_length: 4  
3. visited_flag: True  
## Process
Step 1: Process Message 1. Since `visited_flag` is `False`, proceed:  
- Update `visited` to `True`.  
- Update `path_length`: `max(1, 2) + 1 = 3`.  
- Update `max_path_length`: `max(2, 3) = 3`.  
Step 2: Process Message 2. Since `visited_flag` is `True`, ignore this
message and do not make any changes.
## Output
State:  
1. visited: True  
2. path_length: 3  
3. max_path_length: 3  

\end{lstlisting}
\end{tcolorbox}

\begin{tcolorbox}[colback=gray!10, colframe=black, rounded corners, boxrule=1.5pt, listing only, listing options={basicstyle=\small\ttfamily, breaklines=true, columns=fullflexible, numbers=none, xleftmargin=0pt}, left=2mm, right=2mm, top=1mm, bottom=1mm]
\begin{lstlisting}[numbers=none, xleftmargin=0pt]
### Termination
The algorithm continues until:
- A node reaches a `path_length` equal to the total number of nodes
(indicating a Hamiltonian path exists).
- Or until a predefined number of iterations equal to the total number 
- of nodes (`N`) is reached without discovering a Hamiltonian path.
Master: ##################### ITER 3, STEP 1: REASONING ######################
Master: {'thought': "The algorithm is designed to check for a 
Hamiltonian path. Now let's execute it to see if such a path exists.", 
'speak': "The algorithm is ready. Let's execute it to determine if there is
a Hamiltonian path.", 'function': [{'name': 'Distributed_Execution', 
'arguments': {}}]}
Master: ####################### ITER 3, STEP 2: ACTING #######################
0: State: 1. visited: True 2. path_length: 1 3. max_path_length: 1
2: State: 1. visited: False 2. path_length: 0 3. max_path_length: 1
5: State: 1. visited: False 2. path_length: 0 3. max_path_length: 1
3: State: 1. visited: False 2. path_length: 0 3. max_path_length: 1
4: State: 1. visited: False 2. path_length: 0 3. max_path_length: 1
1: State: 1. visited: False 2. path_length: 0 3. max_path_length: 1
Node 5 Send Message to Node 1: 1. path_length: 1 2. max_path_length: 
1 3. visited_flag: False
Node 5 Send Message to Node 2: 1. path_length: 1 2. max_path_length:
1 3. visited_flag: False
Node 5 Send Message to Node 3: 1. path_length: 1 2. max_path_length:
1 3. visited_flag: False 
...
All agents' state unchanged, terminating early...
>>> END 
system: 1. Execute function Distributed_Execution
   [ARGUMENTS]:
       
   [STATUS]: SUCCESS
   [RESULT]: Node: 0  State: 1. visited: True 2. path_length: 5 3. max_path_length: 1
Node: 1  State: 1. visited: True 2. path_length: 2 3. max_path_length: 1
Node: 2  State: 1. visited: True 2. path_length: 4 3. max_path_length: 3
Node: 3  State: 1. visited: True 2. path_length: 5 3. max_path_length: 3
Node: 4  State: 1. visited: True 2. path_length: 4 3. max_path_length: 3
Node: 5  State: 1. visited: True 2. path_length: 5 3. max_path_length: 3

Master: ##################### ITER 4, STEP 1: REASONING ######################
Master: {'thought': 'Since the maximum path length found is 5, which is
less than the total number of nodes (6), there is no Hamiltonian path in 
the graph.', 'speak': 'No, there is no Hamiltonian path in this graph.', 
'function': []}
\end{lstlisting}
\end{tcolorbox}

\subsection{Other Execution Examples}
\textbf{Shortest Path Execution Example.}

Problem Description: 

Find the shortest distance from a source node to other nodes in an undirected graph. In an undirected graph, (i,j,k) means that node i and node j are connected with an undirected edge with weight k. The graph has 8 nodes, and the edges are: (0,7,9) (0,1,7) (0,4,9) (1,7,1) (2,7,7) (2,6,5) (2,5,8) (3,5,9) (3,4,8) (3,6,1) (4,7,7) (4,5,6) (5,7,6). Give the weight of the shortest distance from node 1 to other node.

Execution Process:

\begin{tcolorbox}[colback=gray!10, colframe=black, rounded corners, boxrule=1.5pt, listing only, listing options={basicstyle=\small\ttfamily, breaklines=true, columns=fullflexible, numbers=none, xleftmargin=0pt}, left=2mm, right=2mm, top=1mm, bottom=1mm]
\begin{lstlisting}[numbers=none, xleftmargin=0pt]
Master: ##################### ITER 1, STEP 1: REASONING ######################
Master: {'thought': 'I need to find the shortest distances from node 1 to all 
other nodes in the given undirected graph. First, I will build the graph using 
the provided edges and then design a distributed algorithm to find the 
shortest paths.', 
\end{lstlisting}
\end{tcolorbox}

\begin{tcolorbox}[colback=gray!10, colframe=black, rounded corners, boxrule=1.5pt, listing only, listing options={basicstyle=\small\ttfamily, breaklines=true, columns=fullflexible, numbers=none, xleftmargin=0pt}, left=2mm, right=2mm, top=1mm, bottom=1mm]
\begin{lstlisting}[numbers=none, xleftmargin=0pt]
'speak': 'Building the graph and designing the distributed algorithm to find the shortest distances from node 1.', 'function': '[{"name": 
"Build_Graph", "arguments": {"Edges": [[0, 7, 9], [0, 1, 7], [0, 4, 9], [1, 7, 1], [2, 
7, 7], [2, 6, 5], [2, 5, 8], [3, 5, 9], [3, 4, 8], [3, 6, 1], [4, 7, 7], [4, 5, 6], [5, 7, 6]], 
"Nodes_Num": 8, "Directed": false}}]'}
Master: ####################### ITER 1, STEP 2: ACTING #######################
>>>     Edges: [[0, 7, 9], [0, 1, 7], [0, 4, 9], [1, 7, 1], [2, 7...
>>>     Nodes_Num: 8
>>>     Directed: False
>>> END 
system: 1. Execute function Build_Graph
   [ARGUMENTS]:
       Edges: [[0, 7, 9], [0, 1, 7], [0, 4, 9], [1, 7, 1], [2, 7, 7], [2, 6, 5], [2, 5, 8], [3, 
       5, 9], [3, 4, 8], [3, 6, 1], [4, 7, 7], [4, 5, 6], [5, 7, 6]]
                Nodes_Num: 8
                Directed: False
   [STATUS]: SUCCESS
Master: ##################### ITER 2, STEP 1: REASONING ######################
Master: {'thought': 'The graph has been successfully built. Now, I will design 
a distributed algorithm to find the shortest distances from node 1 to all 
other nodes.', 'speak': 'Designing a distributed algorithm to find the 
shortest distances from node 1.', 'function': '[{"name": 
"Design_Distributed_Algorithm", "arguments": {"Problem_Str": "Find the 
shortest distance from node 1 to all other nodes in an undirected graph 
with 8 nodes and given edges."}}]'}
Master: ####################### ITER 2, STEP 2: ACTING #######################
>>> END 
Master: ##################### ITER 3, STEP 1: REASONING ######################
Master: {'thought': 'The distributed algorithm has been successfully 
designed. Now, I will execute the algorithm to find the shortest distances
from node 1 to all other nodes.', 'speak': 'Executing the distributed 
algorithm to calculate the shortest distances from node 1.', 'function': 
'[{"name": "Distributed_Execution", "arguments": {}}]'}
Master: ####################### ITER 3, STEP 2: ACTING #######################
Initialization:
1: State: 1. distance: 0
0: State: 1. distance: \infinity
2: State: 1. distance: \infinity
4: State: 1. distance: \infinity
5: State: 1. distance: \infinity
7: State: 1. distance: \infinity
3: State: 1. distance: \infinity
6: State: 1. distance: \infinity
Node 6 Send Message to Node 2: 1. new_distance: \infinity  
Node 6 Send Message to Node 3: 1. new_distance: \infinity
Node 1 Send Message to Node 0: 1. new_distance: 7  
Node 1 Send Message to Node 7: 1. new_distance: 1
Node 3 Send Message to Node 5: 1. new_distance: \infinity  
....
All agents' state unchanged, terminating early...
>>> END 
system: 1. Execute function Distributed_Execution
   [ARGUMENTS]:
       
   [STATUS]: SUCCESS
   [RESULT]: Node: 0  State: 1. distance: 7
Node: 1  State: 1. distance: 0
Node: 2  State: 1. distance: 8
Node: 3  State: 1. distance: 14
Node: 4  State: 1. distance: 8
Node: 5  State: 1. distance: 7
Node: 6  State: 1. distance: 13
Node: 7  State: 1. distance: 1
\end{lstlisting}
\end{tcolorbox}

\begin{tcolorbox}[colback=gray!10, colframe=black, rounded corners, boxrule=1.5pt, listing only, listing options={basicstyle=\small\ttfamily, breaklines=true, columns=fullflexible, numbers=none, xleftmargin=0pt}, left=2mm, right=2mm, top=1mm, bottom=1mm]
\begin{lstlisting}[numbers=none, xleftmargin=0pt]
Master: ##################### ITER 4, STEP 1: REASONING ######################
The shortest distances from node 1 are as follows: Node 0: 7, Node 1: 0, 
Node 2: 8, Node 3: 14, Node 4: 8, Node 5: 7, Node 6: 13, Node 7: 1.Master: 
{'thought': 'The distributed algorithm has been successfully executed, and I
have obtained the shortest distances from node 1 to all other nodes.', 
'speak': 'The shortest distances from node 1 are as follows: Node 0: 7, Node
1: 0, Node 2: 8, Node 3: 14, Node 4: 8, Node 5: 7, Node 6: 13, Node 7: 1.', 
'function': []}    
\end{lstlisting}
\end{tcolorbox}

\subsection{Distributed Algorithms}
\label{app:distributed}
\paragraph{Example of distributed algorithms in distributed algorithm library.}
\label{app:distributed_in}

~\\
\textbf{Shortest Path.} This distributed algorithm computes the shortest distance from a source node to all other nodes in a weighted graph. Each node maintains a state variable \texttt{distance} representing the known shortest distance from the source node. During initialization, the source node sets its \texttt{distance} to 0, while all other nodes set their \texttt{distance} to infinity. In each round, nodes send messages containing \texttt{new\_distance} to their neighbors, where \texttt{new\_distance} is calculated as the sum of the current node's \texttt{distance} and the edge weight to each neighbor. Upon receiving messages, each node updates its \texttt{distance} if the received \texttt{new\_distance} is smaller than its current \texttt{distance}. The algorithm terminates when no node updates its state, ensuring that all nodes have computed their shortest distances from the source. See Figure~\ref{fig:spd}.

\noindent \textbf{Connectivity.} This distributed algorithm determines connected components in a graph by propagating component identifiers. Each node maintains a state variable \texttt{Component\_Id} representing the identifier of the connected component it belongs to. During initialization, each node sets its \texttt{Component\_Id} to its own node ID. In each round, nodes send messages containing their current \texttt{Sender\_Component\_Id} to all neighbors. Upon receiving messages, each node compares the received \texttt{Sender\_Component\_Id} with its current \texttt{Component\_Id}. If the received ID is smaller, the node updates its \texttt{Component\_Id} to the smaller value; otherwise, it keeps its current \texttt{Component\_Id} unchanged. The algorithm terminates when no node updates its state, at which point nodes with the same \texttt{Component\_Id} belong to the same connected component. See Figure~\ref{fig:connect}.

\begin{figure*}[htbp]
\centering
\includegraphics[width=0.8\textwidth]{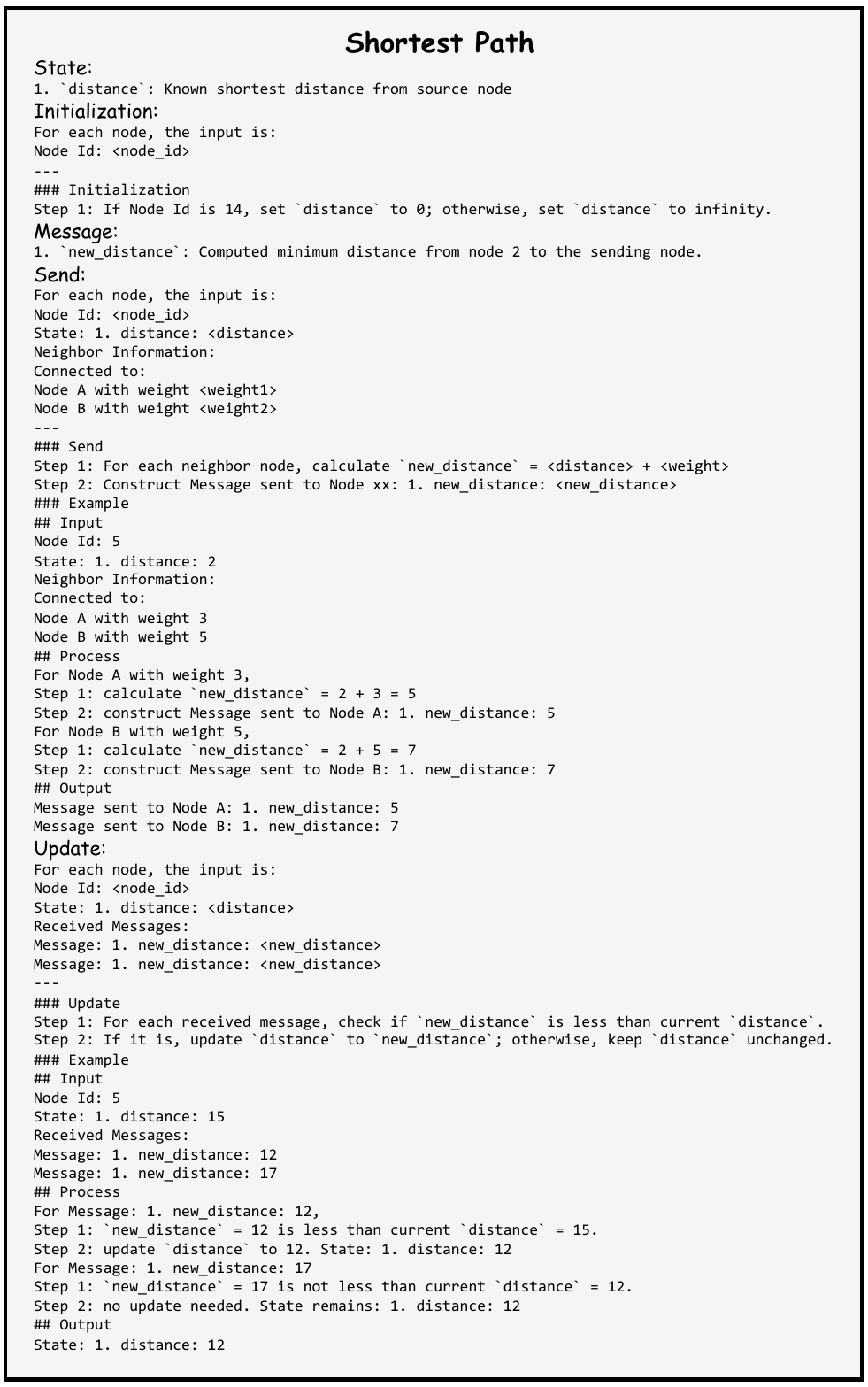}
\caption{Distributed algorithm for shortest path problem under the distributed paradigm.}
\label{fig:spd}
\vspace{-1.2em}
\end{figure*}

\begin{figure*}[htbp]
\centering
\includegraphics[width=0.8\textwidth]{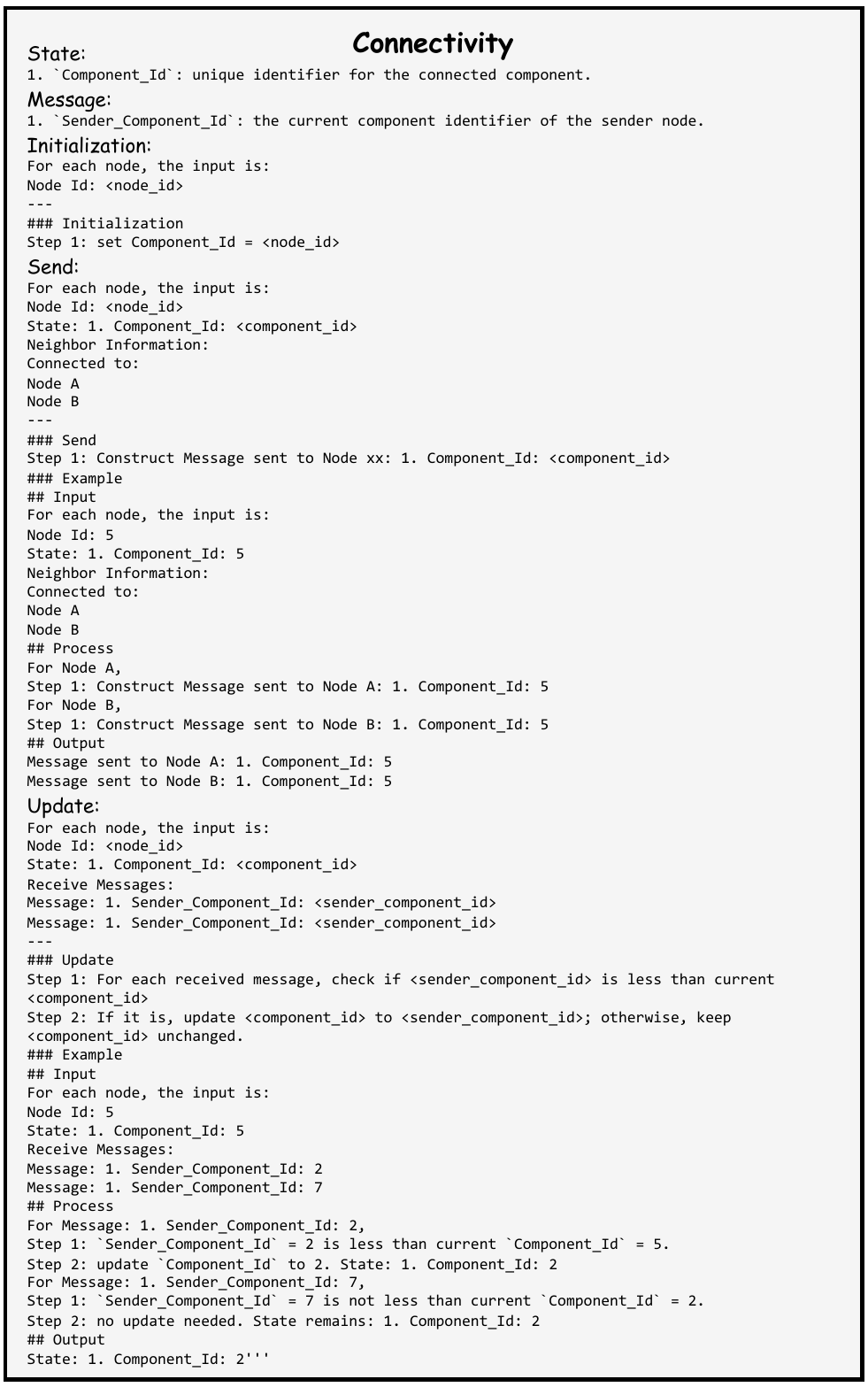}
\caption{Distributed algorithm for connectivity problem under the distributed paradigm.}
\label{fig:connect}
\vspace{-1.2em}
\end{figure*}

\paragraph{Example of distributed algorithms designed by the Master LLM.}
\label{app:distributed_out}
~\\
\textbf{PageRank.} This distributed algorithm computes the importance score (PageRank) of webpages in a directed graph. Each node maintains a state variable \texttt{rank} representing the importance of the webpage, initialized to 1.0. In each round, nodes calculate their \texttt{contribution} as \texttt{rank / out\_degree}, where \texttt{out\_degree} is the total number of outgoing links, and send this \texttt{contribution} to all connected neighbors. Upon receiving messages, each node sums all received \texttt{contribution} values and updates its \texttt{rank} using the PageRank formula: \texttt{new\_rank = damping\_factor * (sum\_of\_received\_contributions) + (1 - damping\_factor) * initial\_rank}, where \texttt{damping\_factor} is typically set to 0.85. The algorithm iterates until the rank values converge, ensuring that each node has computed its final PageRank score. See Figure~\ref{fig:pagerank}.

\noindent \textbf{Hamilton Path.} This distributed algorithm determines whether a Hamiltonian path exists in a graph by propagating path information. Each node maintains two state variables: \texttt{isInPath} (a boolean indicating if the node is part of the Hamiltonian path) and \texttt{pathLength} (an integer representing the length of the path ending at this node). During initialization, a starting node (e.g., Node 0) sets \texttt{isInPath} to \texttt{True} and \texttt{pathLength} to 1, while all other nodes set \texttt{isInPath} to \texttt{False} and \texttt{pathLength} to 1. In each round, nodes with \texttt{isInPath} = \texttt{True} send messages containing their \texttt{pathLength} + 1 and \texttt{isInPath} status to neighbors. Upon receiving messages with \texttt{isInPath} = \texttt{True}, each node updates its \texttt{pathLength} to the maximum of its current value and the received \texttt{pathLength}, and sets \texttt{isInPath} to \texttt{True} if the received path is longer. The algorithm terminates when no updates occur, and a Hamiltonian path exists if any node reaches a \texttt{pathLength} equal to the total number of nodes. See Figure~\ref{fig:hamilton}.

\noindent \textbf{Subgraph Matching.} This distributed algorithm verifies whether there exists a subgraph in graph $\mathcal{G}$ that is isomorphic to a given subgraph $\mathcal{G}'$. Each node in graph $\mathcal{G}$ maintains two state variables: \texttt{NodeMatch} (a list of potential matches for nodes in the subgraph $\mathcal{G}'$) and \texttt{MatchedFlag} (a boolean indicating if the node is currently matched to a node in the subgraph). During initialization, each node in $\mathcal{G}$ sets \texttt{NodeMatch} as an empty list and \texttt{MatchedFlag} as false, while each node in subgraph $\mathcal{G}'$ initializes its unique identifier and sets \texttt{MatchedFlag} as false. In each round, nodes send two types of messages: \texttt{MatchInfo} (containing the node identifier and its list of current potential matches) and \texttt{EdgeConnection} (indicating the directed neighbors that the sending node is connected to). Upon receiving \texttt{MatchInfo} messages, each node checks if the identifier matches any node in subgraph $\mathcal{G}'$, and if a match is found, updates \texttt{NodeMatch} to include the new match and sets \texttt{MatchedFlag} to true. For \texttt{EdgeConnection} messages, nodes update their list of potential neighbors or confirm connectivity with received nodes, ensuring to track relationships needed for further matching. The algorithm terminates when no new matches are found and the \texttt{MatchedFlag} remains false, retaining the current state while preparing for potential future updates. See Figure~\ref{fig:subgraph}.

\begin{figure*}[htbp]
\centering
\includegraphics[width=0.8\textwidth]{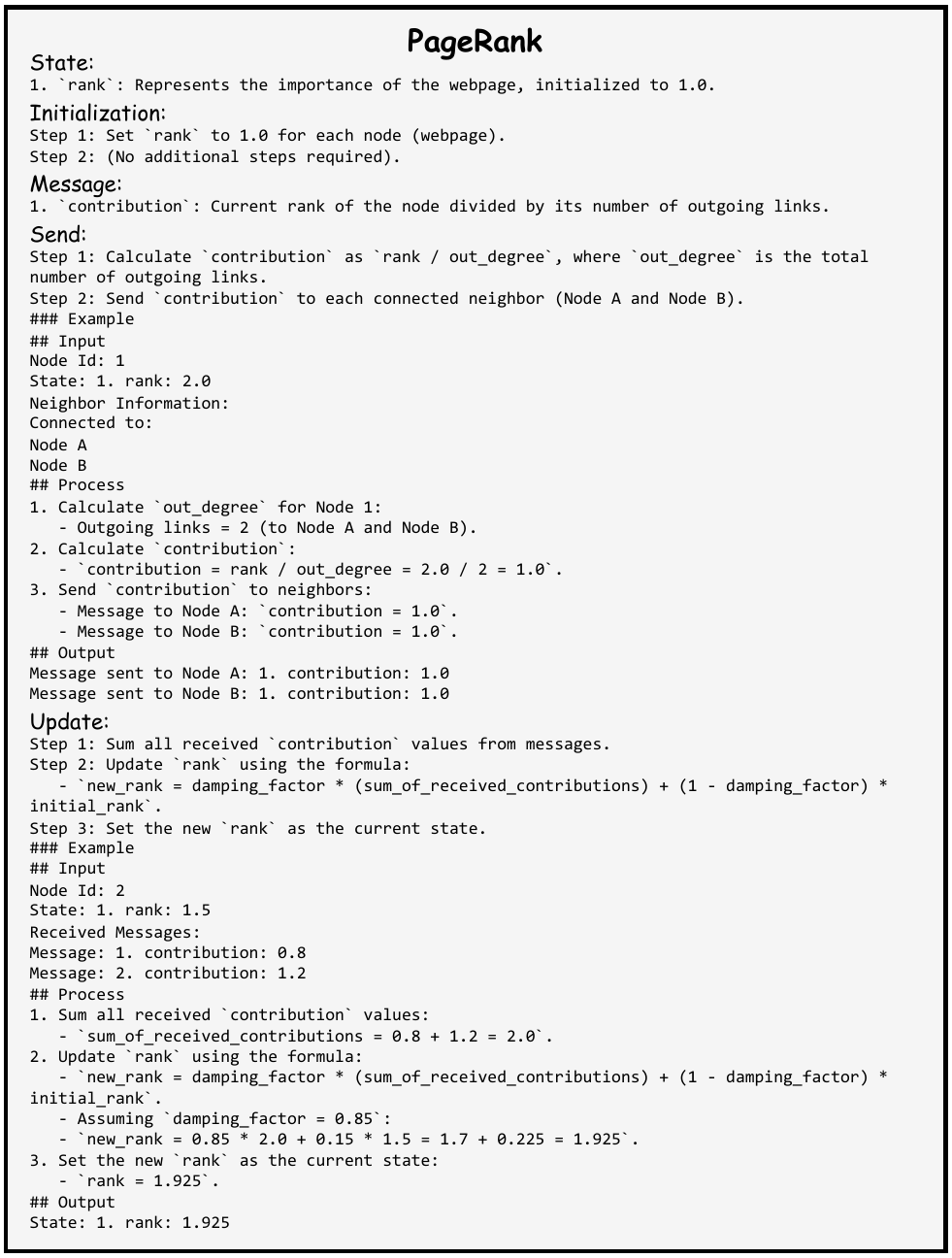}
\caption{Distributed algorithm for pagerank calculation under the distributed paradigm.}
\label{fig:pagerank}
\vspace{-1.2em}
\end{figure*}

\begin{figure*}[htbp]
\centering
\includegraphics[width=0.8\textwidth]{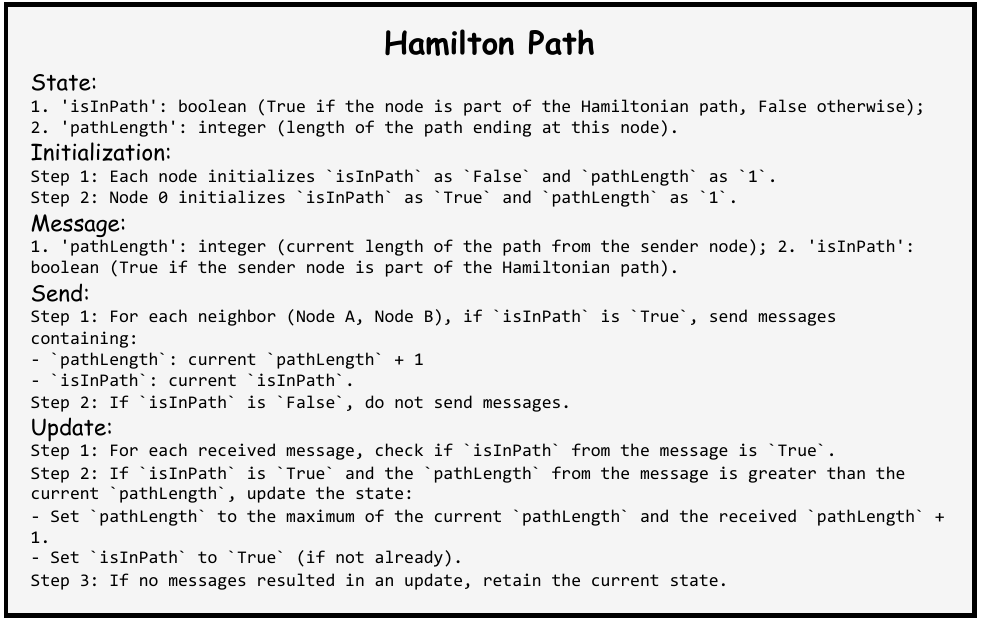}
\caption{Distributed algorithm for hamilton path problem under the distributed paradigm.}
\label{fig:hamilton}
\vspace{-1.2em}
\end{figure*}

\begin{figure*}[htbp]
\centering
\includegraphics[width=0.8\textwidth]{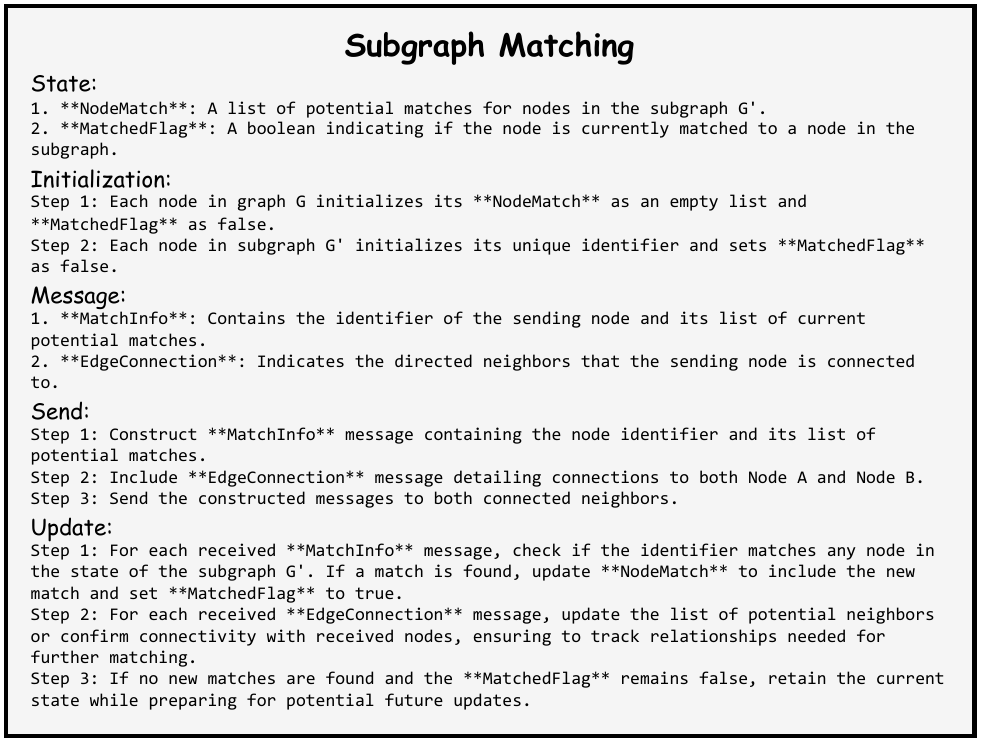}
\caption{Distributed algorithm for subgraph matching problem under the distributed paradigm.}
\label{fig:subgraph}
\vspace{-1.2em}
\end{figure*}

}

\end{document}